\documentclass[11pt]{article}

\usepackage[preprint]{acl}

\usepackage{pifont}
\usepackage{times}
\usepackage{latexsym}
\usepackage[T1]{fontenc}
\usepackage[utf8]{inputenc}
\usepackage{microtype}
\usepackage{inconsolata}
\usepackage{graphicx}
\usepackage{booktabs}
\usepackage{multirow}
\usepackage{amsmath}
\usepackage{amssymb}
\usepackage{fontawesome}
\usepackage{soul}
\usepackage{array}
\usepackage{makecell}
\usepackage{rotating}
\usepackage{booktabs}
\usepackage{arydshln}
\usepackage{tcolorbox}
\tcbuselibrary{breakable}
\usepackage{colortbl}
\usepackage[table,xcdraw]{xcolor}
\usepackage{overpic}
\usepackage{placeins}
\usepackage{subcaption}
\newcommand{\ARGTCA}{\textsc{ArgTca}}
\newcommand{\ARGTCADiv}{\textsc{ArgTca-Div}}
\newcommand{\ARGTCADisc}{\textsc{ArgTca-Disc}}
\newcommand{\TCA}{\textsc{TCA}}
\newcommand{\CTPT}{\textsc{C-TPT}}
\newcommand{\OTPT}{\textsc{O-TPT}}
\newcommand{\TPT}{\textsc{TPT}}
\newcommand{\CLIP}{\textsc{CLIP}}
\newcommand{\GAT}{\textsc{GAT}}

\newcommand{\ECE}{\textsc{ECE}}
\newcommand{\LLM}{\textsc{LLM}}
\newcommand{\VLM}{\textsc{VLM}}
\newcommand{\SAG}{\textsc{Sag}}   

\newcommand{\h}{\mathbf{h}}

\newcommand{\p}{\mathbf{p}}
\newcommand{\W}{\mathbf{W}}

\newcommand{\hL}{\h^{(L)}}
\newcommand{\hzero}{\h^{(0)}}
\newcommand{\R}{\mathbb{R}}
\newcommand{\calE}{\mathcal{E}}

\newcommand{\calN}{\mathcal{N}}
\newcommand{\calA}{\mathcal{A}}

\newcommand{\htilde}{\tilde{\h}}

\usepackage[normalem]{ulem}

\definecolor{bestblue}{RGB}{0,90,180}

\newcommand{\best}[1]{%
  \sethlcolor{red!30}\hl{\textbf{#1}}%
}
\newcommand{\second}[1]{%
  \sethlcolor{orange!30}\hl{#1}%
}

\definecolor{phase1}{RGB}{0,100,180}
\definecolor{eqbox}{RGB}{240,248,255}

\newcommand{\norm}[1]{\left\lVert#1\right\rVert}

\newtcolorbox{eqblock}[1][]{
  colback=eqbox, colframe=phase1!60, fonttitle=\bfseries\small,
  title=#1, breakable, left=6pt, right=6pt, top=4pt, bottom=4pt
}

\title{\ARGTCA{}: Attribute Relation Graph for\\
       Test-Time Calibration of Vision-Language Models}

       \title{When Prompts Ignore Structure: Graph-Based Attribute Reasoning for Calibrated VLMs}

\author{
Tanay Sodha\thanks{\hspace{-0.4em}Equal contribution.}$^{1}$ \and
Aditya Sharma\footnotemark[1]$^{1}$ \and
Ramya Hebbalaguppe$^{2}$ \and
Vinti Agarwal$^{1}$ \\ \and 
\textbf{Pranav Murthy Yeluripaty}$^{1}$ \\
$^{1}$Department of Computer Science and Information Systems,\\
Birla Institute of Technology and Science, Pilani, India \\
$^{2}$TCS Research, New Delhi, India \\
\texttt{}
}

\begin{document}
\maketitle



\begin{abstract}

Reliable confidence estimation remains a key limitation of test-time adaptation in vision--language models (VLMs), where prompt tuning improves zero-shot accuracy but often degrades calibration due to entropy-driven overconfidence. Prior approaches mitigate this using \LLM{}-derived class attributes and contrastive regularization, yet treat attributes independently, ignoring their relational structure. We propose \ARGTCA{}, 
which represents (class, attribute) pairs as nodes in a Symbolic Attribute Graph
and trains a Graph Attention Network (GAT) via contrastive objectives to produce structurally informed embeddings capturing inter-attribute dependencies.

We introduce two attribute selection strategies: \ARGTCADiv{} for intra-class diversity and \ARGTCADisc{} for inter-class discrimination.   
Experiments across $9$ benchmarks show that \ARGTCADiv{} reduces average ECE($\downarrow$) by $\sim37\%$ over baselines, 
whereas \ARGTCADisc{} consistently performs as the second-best variant, reducing average ECE by $\sim17\%$ over baselines. These results suggest that modeling symbolic attribute interactions provides a principled approach for reliable test-time adaptation in VLMs. 
\end{abstract}


\section{Introduction}
\label{sec:intro}

Vision-language models (\VLM{}s) such as \CLIP{} \citep{radford2021learning} have shown strong zero-shot image recognition by aligning images and text in a shared embedding space through large-scale contrastive pretraining. \CLIP{} performs zero-shot classification of an image by computing cosine similarities between its visual embedding and class-conditioned text features generated from prompt templates such as \emph{``a photo of a \{class\}''}.  While effective, such templates are suboptimal and domain-agnostic. 
Test-time prompt tuning (\TPT{}) \citep{shu2022test} addresses this by optimizing prompt tokens for each test image using only its own augmented views, with no labeled data. However, \TPT{}s entropy minimization objective inherently drives the model toward overconfident predictions, producing miscalibrated outputs, quantified by the Expected
Calibration Error (\ECE{}) \citep{guo2017calibration}.  It poses a fundamental barrier to deploying \VLM{}s in  safety-sensitive applications such as healthcare diagnostics and autonomous systems, where unreliable uncertainty estimates can have serious consequences. 

\begin{figure}[t]
    \centering
    \setlength{\tabcolsep}{1pt}
    \begin{tabular}{ccc}
        \includegraphics[width=0.30\linewidth]{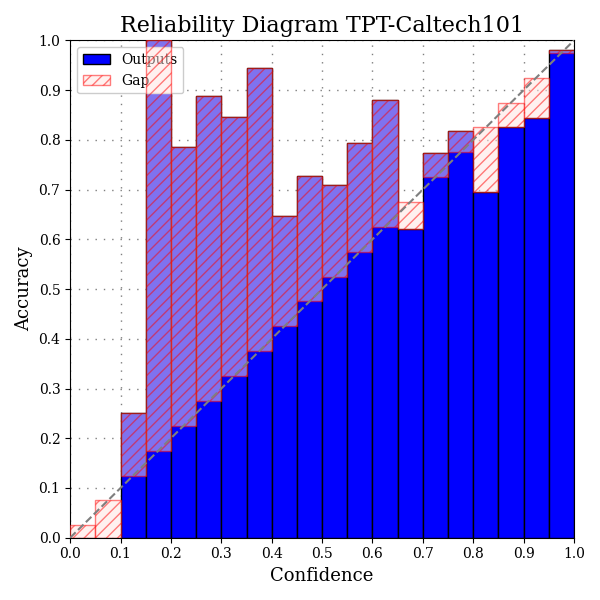} &
        \includegraphics[width=0.30\linewidth]{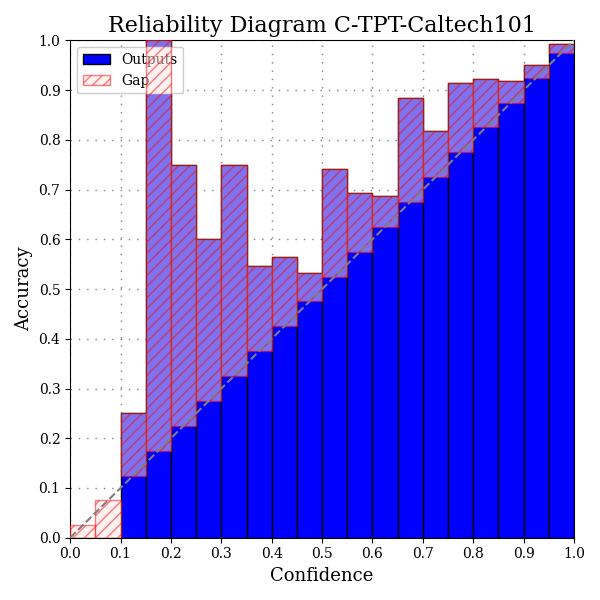} &
        \includegraphics[width=0.30\linewidth]{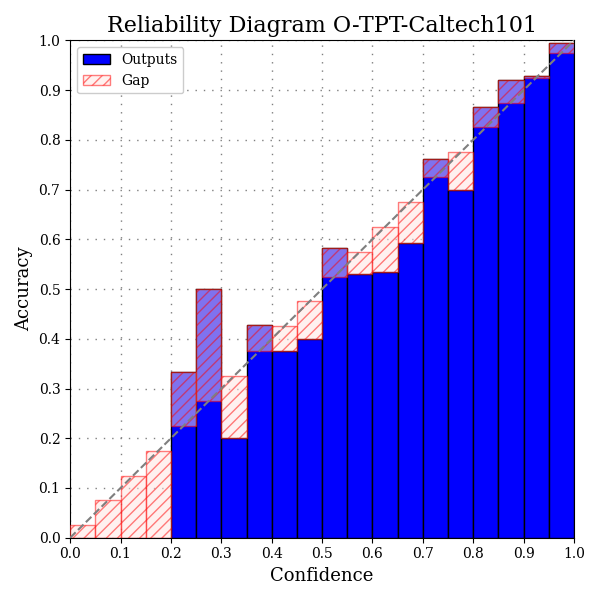} \\[1pt]
        {\small TPT (4.59)} &
        {\small \CTPT{} (4.24)} &
        {\small \TCA{} (2.56)} \\[4pt]
        \includegraphics[width=0.30\linewidth]{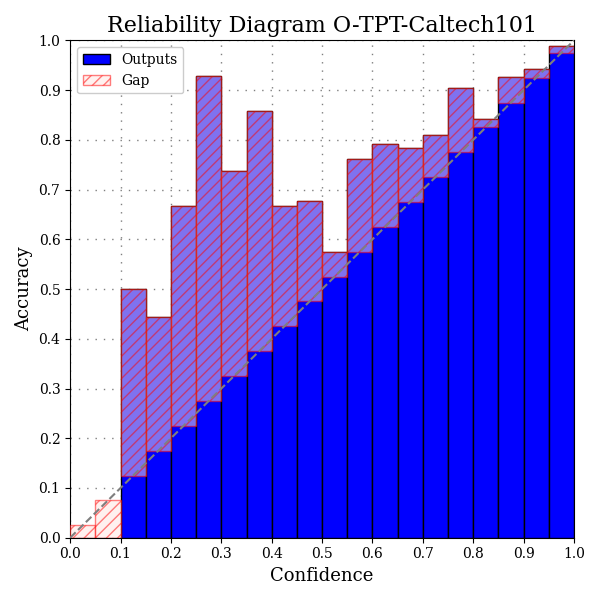} &
        \includegraphics[width=0.30\linewidth]{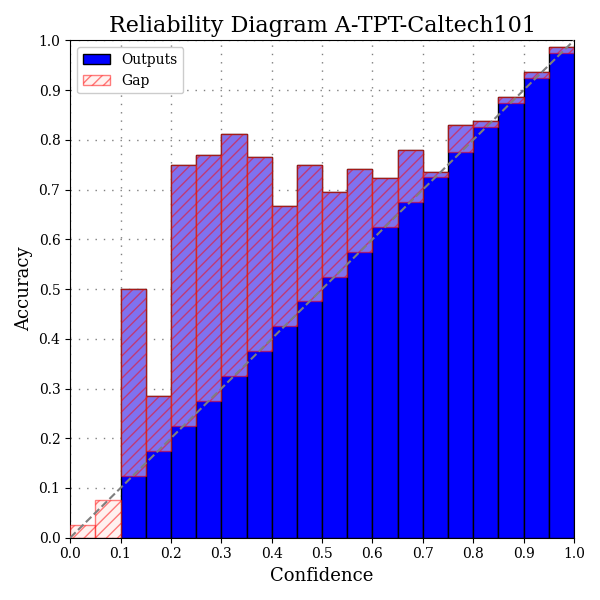} &
        \includegraphics[width=0.30\linewidth]{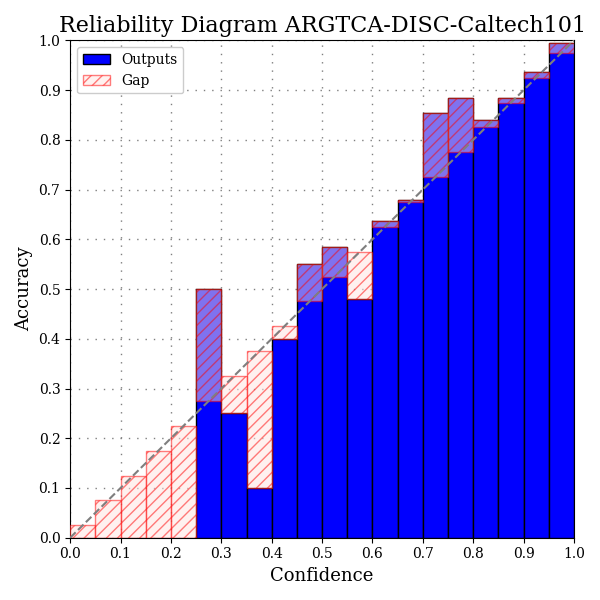} \\[1pt]
        {\small \OTPT{} (3.62)} &
        {\small A-TPT (2.67)} &
        {\small \textbf{Ours} (1.62)} \\
    \end{tabular}
    \caption{Reliability diagrams on Caltech101 dataset (ViT-B/16, \ECE{}($\downarrow$) in parentheses). \TCA{} is overconfident (bars below diagonal); while TPT, \CTPT{}, \OTPT{},
and A-TPT overcorrect into underconfidence (bars above diagonal).
 \ARGTCADisc{} (ours) achieves the closest alignment with the calibration diagonal across all confidence bins. Full results across all datasets are provided in Figure~\ref{fig:reliability-all} in Appendix.}
    \label{fig:reliability-caltech}
\end{figure}
A key insight motivating recent work is that calibration and accuracy are largely \emph{decoupled}: different prompts can achieve nearly identical top-1 accuracy
while exhibiting significantly divergent \ECE{} \citep{yoon2024ctpt}. This implies that the \textit{geometry of the textual feature space}, rather than predictive accuracy alone, governs how well-calibrated a \VLM{}s predictions are. \CTPT{} \citep{yoon2024ctpt} and \OTPT{}\cite{sharifdeen2025otpt} exploit this by adding a feature-dispersion loss and angular dispersion respectively, that spreads class-conditioned text features across the hypersphere. \TCA{} \citep{hebbalaguppe2025tca} complements these geometric objectives by adding \LLM{}-extracted visual attributes for semantically grounded prompt initialization, combined with a contrastive intra-class and inter-class regularization over the resulting text embeddings, achieving strong calibration.

\paragraph{Structural limitations of \TCA{}.}
Despite strong calibration performance, \TCA{} exhibits two structural failures that directly degrade the geometry of class-conditioned text features. First, attributes are selected as a flat set based on cosine-similarity without any relational context: on texture-rich datasets such as DTD, attributes such as \textit{rough}, \textit{bumpy}, and \textit{grainy} are nearly collinear in \CLIP{'s}
embedding space, so top-$M'$ selection clusters attribute embeddings tightly and undermines the separation that calibration requires. Second, attributes shared ubiquitously across different classes---such as \textit{mammal} across \textit{leopard}, \textit{dolphin}, and \textit{beaver} in Caltech101---carry diminished discriminative capacity, leaving their embeddings proximate across class boundaries. Both the failure modes are invisible to \TCA{'s} selection procedure, which operates purely on similarity based ordering of LLM generated attributes without any notion of relational geometry between attributes. 

We propose \ARGTCA{} (\textbf{A}ttribute \textbf{R}elation \textbf{G}raph for \textbf{T}est-time \textbf{C}alibration \textbf{A}daptation), a graph-based framework that addresses both structural failures through a Symbolic Attribute Graph (SAG) constructed by representing (class, attribute) pairs as nodes, trained offline via a Graph Attention Network with no task supervision. Two selection strategies derived from the graph — \ARGTCADiv{}{} for intra-class diversity and \ARGTCADisc{}{} for inter-class discrimination. 
Both strategies operate at zero additional test-time cost — \textbf{Note:} the graph is trained once offline and selection is performed before tuning begins.

We validate \ARGTCA{} across nine benchmark datasets spanning fine-grained recognition, texture classification, and domain-shifted imagery 
and demonstrate empirically that graph-structured relational reasoning over attributes is a key driver of calibration improvement --- as shown in Figure~\ref{fig:reliability-caltech}, \ARGTCA{} produces reliability diagrams most closely aligned with the perfect calibration diagonal, outperforming existing prompt-tuning baselines.
Our key contributions are:
\begin{itemize}

\item We identify two structural failures in \TCA{}: near-synonymous intra-class
attributes that cluster on the hypersphere, and ubiquitous cross-class attributes that carry no discriminative capacity yet receive no suppression. We construct a Symbolic Attribute Graph~(SAG) from class specific attributes and train a Graph Attention Network via supervised contrastive learning to produce relational embeddings capturing intra-class complementarity and cross-class redundancy.

\item We introduce two selection strategies from graph-refined embeddings:
\ARGTCADiv{}, maximising intra-class semantic complementarity; and \ARGTCADisc{}, maximising distance from all other-class embeddings. Both strategies
operate at zero additional test-time cost and are evaluated across nine benchmarks using CLIP ViT-B/16.

\end{itemize}

\FloatBarrier
\section{Related Work}
\label{sec:related}

\subsection{Test-Time Adaptation of VLMs}
Foundation models such as \CLIP{}~\citep{radford2021learning, jia2021scaling} enable
zero-shot classification by aligning image and text in a shared embedding space.
While learned prompt methods like CoOp~\citep{zhou2022learning} and
CoCoOp~\citep{zhou2022conditional} improve over hard prompts, they require
labeled data, limiting zero-shot applicability. \TPT{}~\citep{shu2022test}
removes this constraint by adapting prompts via entropy minimization over
augmented views of a single test image, extended further by
DiffTPT~\citep{feng2023diverse} and
PromptAlign~\citep{mirza2024promptalign} through improved augmentation and
distribution alignment. However, entropy minimization drives overconfident
predictions across all \TPT{}-family methods, significantly degrading
calibration relative to the zero-shot baseline\cite{yoon2024ctpt}.
\subsection{Calibration of Vision-Language Models}
Post-hoc methods such as temperature scaling~\citep{guo2017calibration} and
Platt scaling~\citep{platt1999probabilistic} require labeled validation data,
making them impractical at test time. \CTPT{}~\citep{yoon2024ctpt} establishes
that \TPT{} degrades calibration, and that well-calibrated prompts exhibit
higher Average Text Feature Dispersion~(ATFD). \OTPT{}~\citep{sharifdeen2025otpt}
refines this, showing that ATFD displaces the feature centroid without ensuring
pairwise angular separation --- the operative quantity for calibration on the
unit hypersphere. A-TPT~\citep{atpt} addresses a further degeneracy in \OTPT{} when $N > |D|$ ($N$: \#datapoints, $D$: feature dimensions), enforcing pairwise angular separation without dimensional constraints. \TCA{}~\citep{hebbalaguppe2025tca} initializes prompts with
\LLM{}-extracted visual attributes and applies intra- and inter-class contrastive regularization, but treats attributes as a flat set ignoring relationships --- a gap our work addresses.


\subsection{Graph-Based VLM Adaptation}
GNNs~\citep{kipf2017semi, velickovic2018graph} have proven effective for learning over relational data. In \VLM{} adaptation, GraphAdapter~\citep{li2023graphadapter}
uses dual knowledge graphs for supervised few-shot transfer, HGCLIP~\citep{zheng2023hgclip} encodes label taxonomies for hierarchical classification, and VCGPrompt~\citep{zhou2024vcgprompt} builds visual concept graphs for prompt learning. These works demonstrate that relational structure over the label space improves representation quality in VLMs, however are designed to operate in supervised settings, not for zero-shot calibration. In contrast, \ARGTCA{} utilizes symbolic graph edges in a fully label-free setting, specifically targeting calibration improvement. 

\uline{\ARGTCA{} occupies the intersection of all three research threads} retaining \TCA{'s} semantically grounded prompt initialization, corrects its flat-set attribute selection through a Symbolic Attribute Graph, and directly optimizes the geometric properties of text feature space that govern calibration — none of the prior works simultaneously achieves.
\FloatBarrier
\section{Methodology}
\label{sec:method}

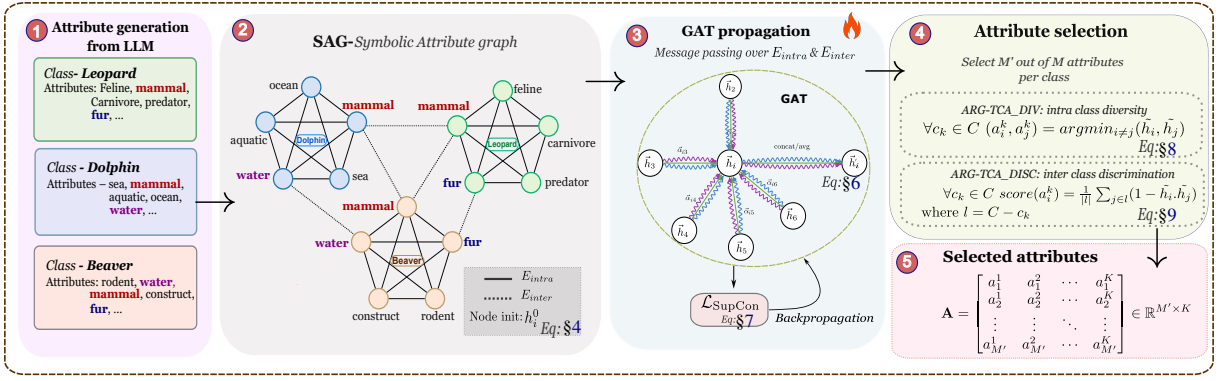
\begin{figure*}
    \centering
    \includegraphics[width=\linewidth]{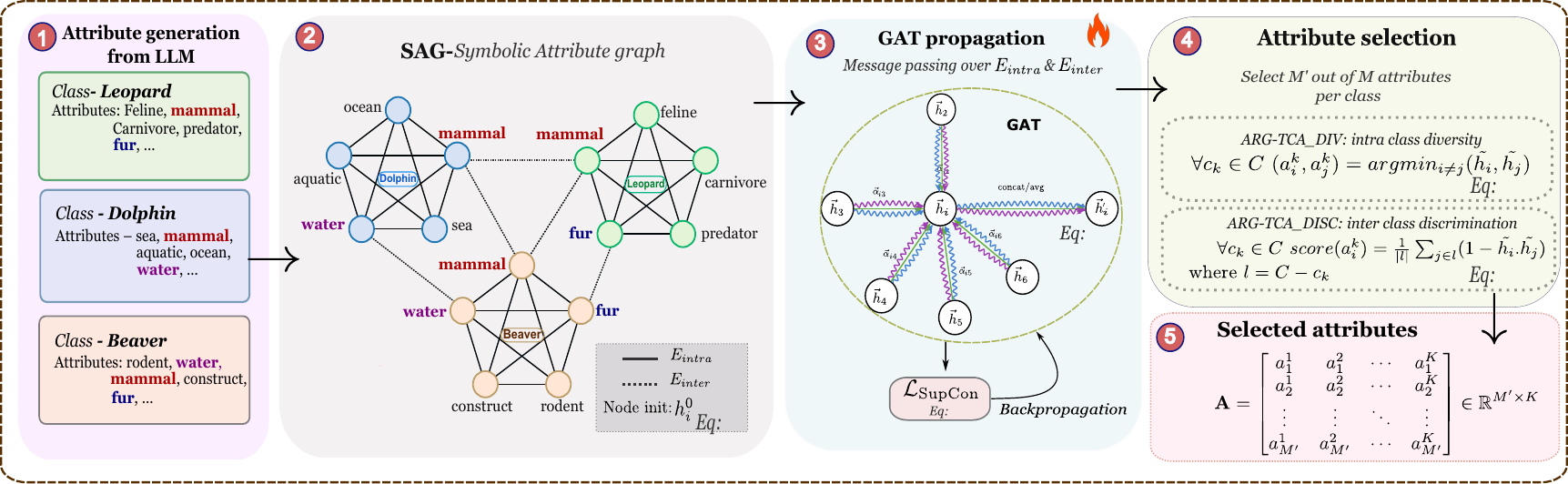}
    \put(-245,16){\scriptsize{\S \ref{eq:h0}}}
    \put(-180,18){\scriptsize{\S \ref{eq:supcon}}}
    \put(-140,70){\scriptsize{\S \ref{eq:gat}}}
    \put(-21,83){\scriptsize{\S \ref{eq:div}}}
    \put(-21,58){\scriptsize{\S \ref{eq:disc}}}


\caption{Overview of \ARGTCA{} attribute selection pipeline. 
\ding{182} An LLM generates $M$ candidate attributes per class. 
\ding{183} These are used to construct the Symbolic Attribute Graph (SAG) 
with intra-class ($E_{\text{intra}}$) and inter-class ($E_{\text{inter}}$) edges. 
\ding{184} A Graph Attention Network (GAT) is trained via supervised 
contrastive loss ($\mathcal{L}_{\text{SupCon}}$) to propagate structural 
relationships across the SAG, producing attribute embeddings. 
\ding{185} Two selection criteria --- \ARGTCADiv{} (intra-class diversity) 
and \ARGTCADisc{} (inter-class discrimination) --- select the most 
informative $M'$ attributes per class. 
\ding{186} The final selected attributes form the matrix 
$\mathbf{A} \in \mathbb{R}^{M' \times K}$, used for downstream prompt tuning.}

\label{fig:pipeline}
\end{figure*}
\ARGTCA{} replaces \TCA{}'s~\cite{hebbalaguppe2025tca} similarity-based attribute selection with a graph-informed selection via two phases. In \textbf{Phase~1}, a Symbolic 
Attribute Graph~(SAG) is constructed over all $(\text{class, attribute})$ 
pairs, with intra-class edges ($E_{\text{intra}}$) encoding within-class 
attribute relations and inter-class edges ($E_{\text{inter}}$) encoding 
cross-class sharing. A Graph Attention Network~(GAT) is trained with a 
supervised contrastive objective~(Eq.~\ref{eq:supcon}), producing relational 
embeddings that capture \textit{intra}-class complementarity and 
\textit{inter}-class redundancy. These drive two principled selection 
strategies: \ARGTCADiv{}, which selects the most semantically 
complementary attribute pair per class, spanning the widest semantic range 
for diverse prompt adaptation signals; and \ARGTCADisc{}{}, which 
scores each attribute by its mean angular distance from all other-class nodes 
and selects the top-$M'$ per class --- these maximally class-specific 
descriptors strengthen \TCA{}'s inter-class regularizer. Figure~\ref{fig:pipeline} 
gives an overview of the proposed graph-informed attribute selection. \\
In \textbf{Phase~2}, selected 
attributes initialize the standard \TCA{} test-time tuning procedure without modification. \textit{The graph's contribution is purely structural: deciding which attributes appear, not how they are represented.} Moreover, graph training and attribute selection are performed once offline before testing, incurring no additional test-time cost.

\subsection{Preliminaries}
\label{sec:method:prelim}
\paragraph{Notation} Let ${C} = \{c_1,\ldots,c_K\}$ be a set of $K$ class labels and $A_k = \{a_1,\ldots,a_M\}$ be the set of $M$ \LLM{}-generated visual attributes for class $k$, giving $N = K \times M$ total ($c_k,a_j$) pairs.  
A full list of notations is provided in Table \ref{tab:notations} in Appendix.

\paragraph{\TCA{} background}
\TCA{} \citep{hebbalaguppe2025tca} initializes soft prompts using visual
attributes and defines per-class text features via attribute-conditioned prompt embeddings.
For each class $k$ and each of its $M'$ selected attributes, a
prompt is assembled as
\begin{equation}
  t_{k,m} = \mathbf{p} \oplus \mathbf{a}_m \oplus \mathbf{c}_k,
  \label{eq:prompt}
\end{equation}
where $\mathbf{p}$ is the learnable soft context (initialised to
\texttt{"a photo of a"}), and $\mathbf{a}_m$, $\mathbf{c}_k$ are the token embeddings of the attribute and class name strings, respectively. 
The text feature is $f_{k,m} = g(t_{k,m})$, where $g(\cdot)$ denotes the frozen CLIP text encoder. Class probabilities combine all attribute prompts via a bag-of-attributes softmax aggregation:

\begin{equation}
\small
  p(y = k \mid x)
  = \frac{\displaystyle\sum_{m=1}^{M'} \exp\!\bigl( f_{\mathrm{img}}(x)\cdot f_{k,m}/\tau\bigr)}
         {\displaystyle\sum_{k'=1}^{K}\sum_{m=1}^{M'} \exp\!\bigl( f_{\mathrm{img}}(x)\cdot f_{k',m}/\tau\bigr)}
  \label{eq:tca_prob}
\end{equation}
where $\tau$ is CLIP's learned logit scale. The test-time objective minimizes:
\begin{equation}
  \mathcal{L}_{\mathrm{TCA}} = \mathcal{L}_{\mathrm{TPT}}
         - \alpha\,\mathcal{L}_{\mathrm{inter}}
         + \beta\,\mathcal{L}_{\mathrm{intra}}
  \label{eq:tca_loss}
\end{equation}
with $\alpha$ weights how aggressively class centroids $\mu_k$ are pushed away from the global centroid $\mu$, and $\beta$ weights how tightly each class's attribute embeddings $f_{k,m}$ are pulled toward their class centroid $\mu_k$. $\mathcal{L}_{\mathrm{TPT}}$ is the
entropy of the average prediction over the $\lfloor B \cdot p_{\mathrm{sel}}\rfloor$ most confident of $B$ augmented views.
$\mathcal{L}_{\mathrm{inter}} = \frac{1}{K}\sum_k\norm{\mu_k - \mu}_2$ is the
mean distance of class centroids $\mu_k = \frac{1}{M'}\sum_m f_{k,m}$ from
the global centroid $\mu$; maximizing it spreads classes apart.
$\mathcal{L}_{\mathrm{intra}} = \frac{1}{K}\sum_k\frac{1}{M'}\sum_m\norm{f_{k,m}-\mu_k}_2$
is minimized to make each class's attribute prompts consistent.
\emph{Only $\mathbf{p}$ is updated; all other parameters are frozen.}

\subsection{Phase 1 --- Symbolic Attribute Graph and GAT Training}
\label{sec:method:phase1}

\subsubsection{Node Initialization}
\label{sec:method:nodes}

We adopt the $M$ visual attributes per class provided by \citet{hebbalaguppe2025tca}, generated offline via GPT-4 prompting across all nine benchmark datasets.
Specifically, for each class $c_k$, an \LLM{} is prompted with the class name to produce a list of descriptive visual attributes $a_m$, which are subsequently ranked in descending order of relevance by cosine similarity between the attribute and class name embeddings~\citep{hebbalaguppe2025tca}.
Each  class-attribute pair ($c_k$, $a_j$) is represented as a graph node,  initialized with $\ell_2$-normalized EOS hidden state obtained by encoding the phrase ``\texttt{a} $a_{j}$ \texttt{of a} $c_k$'' through frozen \CLIP{} text encoder,$g(\cdot)$:
\begin{equation}
  \hzero_i \;=\; g\!\left(\texttt{"a } a_j \texttt{ of a } c_k\texttt{"}\right)
  \;\in\; \R^{d}
  \label{eq:h0}
\end{equation}
\subsubsection{Symbolic Attribute Graph (\SAG{})}
\label{sec:method:sag}
We define the edge set $E$ via a symbolic rule over the node vocabulary:
\begin{equation}
  \calE = \Bigl\{(i,j) \Big| i \neq j \text{ and } 
  \underbrace{c_i = c_j}_{{E_{intra}}} \text{ or } \underbrace{a_i = a_j}_{{E_{inter}}}\Bigr\}
  \label{eq:edges}
\end{equation}

where both directions are included and self-loops are omitted. In equation \ref{eq:edges} $\calE_{\mathrm{intra}}$ connects all attribute nodes belonging to the same class ($c_i = c_j$), enabling the \GAT{} to jointly aggregate every attribute perspective of a class; $\calE_{\mathrm{inter}}$ connects nodes that sharing a common
attribute token across different classes --- for instance, \emph{mammal}
links \emph{leopard}, \emph{dolphin}, and \emph{beaver} in Caltech101 ---
allowing the GAT to propagate how the same descriptor is contextualized
differently across class neighborhoods.

For $K$ classes each with $M$ attributes, $|\calE_{\mathrm{intra}}| = K \cdot M \cdot (M - 1)$ is guaranteed regardless of the data distribution.

\subsubsection{GAT \& hybrid Objective}
\label{sec:method:supcon}
We use \GAT{} to propagate feature information across $\mathcal{E}_\text{intra}$ and $\mathcal{E}_\text{inter}$ edges so that each attribute node aggregates both the complementary perspectives of its own class and the cross-class context of any shared attribute, producing node embeddings $\hL$ where same-class attributes are pulled into a compact, coherent cluster while cross-class attribute representations are pushed apart.
Concretely, at layer $l$, the updated embedding of node $i$ is computed as:
\begin{equation}
  \h^{\text{l}}_i
    = ELU \left(\Big\|_{h=1}^{H} \sum_{j \in \calN(i)}
          \alpha^h_{j} \cdot \W_h \h_j^{l-1}\right)
  \label{eq:gat}
\end{equation}
where $\|$ denotes head-wise concatenation over $H$ attention heads,
$\mathcal{N}(i)$ is the set of neighbors of node $i$ under
$\mathcal{E}$, and $\mathbf{W}^{(l)}_h \in \mathbb{R}^{d/H \times d}$
is the head-specific linear projection at layer $l$. The attention
coefficient $\alpha^{(l,h)}_{ji}$ measures the relative importance of
neighbor $j$ to node $i$ under head $h$, and $ELU$ induces non-linearity. Full architectural details including the attention coefficient computation are given in section~\ref{app:architecture} of appendix.


\textbf{GAT Training Objective} The GAT is trained using supervised contrastive loss \citep{khosla2020supervised} where same-class $(c_k, a_j)$ nodes form
positives, and all inter-class nodes form negatives.

\begin{equation}
\small
   \mathcal{L}_\text{SupCon} = \frac{1}{|\mathcal{A}|}\sum_{i \in \mathcal{A}} \frac{-1}{|P(i)|} \sum_{p \in P(i)} \\
   \log \frac{\exp\left(\tilde{\mathbf{h}}_i^{\top} \cdot \tilde{\mathbf{h}}_p / \tau\right)}{\displaystyle\sum_{j \neq i} \exp\left(\tilde{\mathbf{h}}_i^{\top} \cdot \tilde{\mathbf{h}}_j / \tau\right)}
    \label{eq:supcon}
\end{equation}
where $\htilde^{(L)} = \hL / \|\hL\|_2$ and
$\calA$ = nodes that have at least one positive.
This trains the \GAT{} to produce class-cohesive representations in $\hL$ space while discriminating against all other classes.

\subsection{Attribute Selection (Post training)}
\label{sec:method:selection}

After GAT training, the learned $\hL$ is used to select $M'$ of the $M$ attributes per class using one of two criteria. Both strategies operate
entirely offline at zero test-time cost.

\paragraph{\ARGTCADiv{} --- Intra-class diversity}
For each class $c_k\in C$, select the pair $(a_i^k, a_j^k)$ with the minimum cosine similarity in the graph-refined embedding space — i.e., the most semantically complementary pair:
\begin{equation}
  (a_i^k, a_j^k) = \arg\min_{i \neq j \in I_k}\;
    \tilde{h}^{(L)}_i \cdot \tilde{h}^{(L)}_j.
  \label{eq:div}
\end{equation}

\paragraph{\ARGTCADisc{}:}
For each node $a_i^k$ in class $c_k$, we compute discriminability score measuring its mean angular distance from all other-class nodes:

\begin{equation}
  \text{score}(a_i^k)
  \;=\; \frac{1}{|\bar{\Omega}_k|}
        \sum_{j \notin \Omega_k}
        \Bigl(1 - \cos\!\bigl(\tilde{h}^{(L)}_i,\;\tilde{h}^{(L)}_j\bigr)\Bigr),
  \label{eq:disc}
\end{equation}
where $\Omega_k$ denotes the global node indices for class $k$ and $\bar{\Omega}_k$ all other-class indices. The top $M'$ attributes by score are selected per class. 

\subsection{Phase 2 --- Test-Time Tuning}
\label{sec:method:phase2}

\paragraph{Prompt construction}
For each $c_k$ and its $M'$ attributes selected in Phase~1, prompts 
are assembled via Eq.~\ref{eq:prompt}, with $\mathbf{a}_m$ and 
$\mathbf{c}_k$ as frozen CLIP token embeddings. The per-class text 
feature is $f_{k,m} = g(t_{k,m})$ and class probabilities follow 
Eq.~\ref{eq:tca_prob}. The full positional breakdown is given in 
Table~\ref{tab:prompt_breakdown}.
\begin{equation}
\small
\label{prompt_layout}
    \text{prompt}\, t_{k,m} =
  \underbrace{[\mathrm{SOS}]}_{\text{pos }0}
  \Big\|
  \underbrace{\mathbf{p}}_{\text{pos }1\text{-}n_\text{ctx}}
  \Big\|
  \underbrace{{a}_m^k, \text{cls tokens},\ldots}_{\text{ pos }n_\text{ctx}+1\;\cdots\;76}
\end{equation}

\paragraph{Test-time objective \& inference}
For each test image, the top $\lfloor B \cdot p_\text{sel} \rfloor$ 
most confident of $B$ augmented views are selected and a single AdamW 
step minimizes $\mathcal{L}_{\text{TCA}}$ (Eq.~\ref{eq:tca_loss}) on 
$\mathbf{p}$. Final prediction uses the non-augmented image with the updated $\mathbf{p}$, which is reset before each new test image. 
Figure~\ref{fig:tca_pipeline} in appendix illustrates the prompt-tuning process.


\FloatBarrier
\section{Experimental Setup}
\label{sec:experiments}

\subsection{Datasets and Baselines}
We evaluate on nine benchmarks datasets, using ViT-B/16 as CLIP backbone and  following the evaluation protocol of baselines~\citep{he2016deep,radford2021learning}. Dataset details are provided in section~\ref{datasets} in appendix. Furthermore, we compare our proposed approach against recent, state-of-the-art approaches:
\begin{itemize}
    \item \textbf{TCA}~\citep{hebbalaguppe2025tca}: The direct
    predecessor, which initializes prompts with LLM-generated visual
    attributes and applies contrastive intra- and inter-class     regularization. Our method builds on and improves upon \TCA{'s}
    attribute selection procedure.
    \item \textbf{C-TPT}~\citep{yoon2024ctpt}: Adds a text feature dispersion loss (ATFD) to TPT~\cite{shu2022test} to improve calibration.
    \item \textbf{O-TPT}~\citep{sharifdeen2025otpt}: Replaces ATFD with pairwise angular separation between class centroids.
    \item \textbf{A-TPT}~\citep{atpt}: Fixes O-TPT's degeneracy when classes exceed embedding dimension.
\end{itemize}  
All baselines are re-evaluated under identical hyperparameters tuned on the Caltech101 validation set, following the protocol of \TCA{} (\citet{hebbalaguppe2025tca}). 

\subsection{Implementation Details}
All experiments are done on a single A100 80GB GPU. The CLIP backbone is fully frozen throughout both phases.
\paragraph{Phase 1 --- Offline Graph Training.}
Each dataset's attribute graph is constructed from $N_{\text{attr}} = 10$
\LLM{}-generated attributes per class (from \citet{hebbalaguppe2025tca}),
giving $K \times 10$ graph nodes.  The Graph Attention Network
(AttributeGAT; 2 layers, 4 heads, 512-dimensional node features) is trained for 100 epochs with supervised contrastive loss (Eq.~\ref{eq:supcon}), temperature $\tau = 0.07$, and the Adam optimizer with learning rate $10^{-3}$.
Phase-1 training is done offline and completes in under two minutes per dataset.

\paragraph{Phase 2 --- Test-Time Prompt Tuning.}
We follow the standard \TCA{} evaluation protocol: soft context initialised to
\emph{``a photo of a''} ($n_{\text{ctx}} = 4$), 64 augmented views per test image with the top 10\% selected by entropy, and a single gradient step per image with batch size 64. Regularization weights: ViT-B/16 uses $\alpha = 10$, $\beta = 35$; 
$M'{=}2$ attributes per class are selected from the graph-refined embeddings via either the diversity (\ARGTCADiv{}) or discriminability (\ARGTCADisc{}) criterion.

\paragraph{Metrics}
We report top-1 accuracy (Acc, \%, $\uparrow$) and Expected
Calibration Error (ECE, \%, $\downarrow$) computed with 20
equal-width bins following~\citet{hebbalaguppe2025tca}. Lower ECE indicates better-calibrated predictions. 

\FloatBarrier

\section{Results}
\label{sec:results}

\subsection{Main Results}
\label{sec:results:main}
\begin{figure}[t]
    \centering
     \includegraphics[width=1\linewidth]{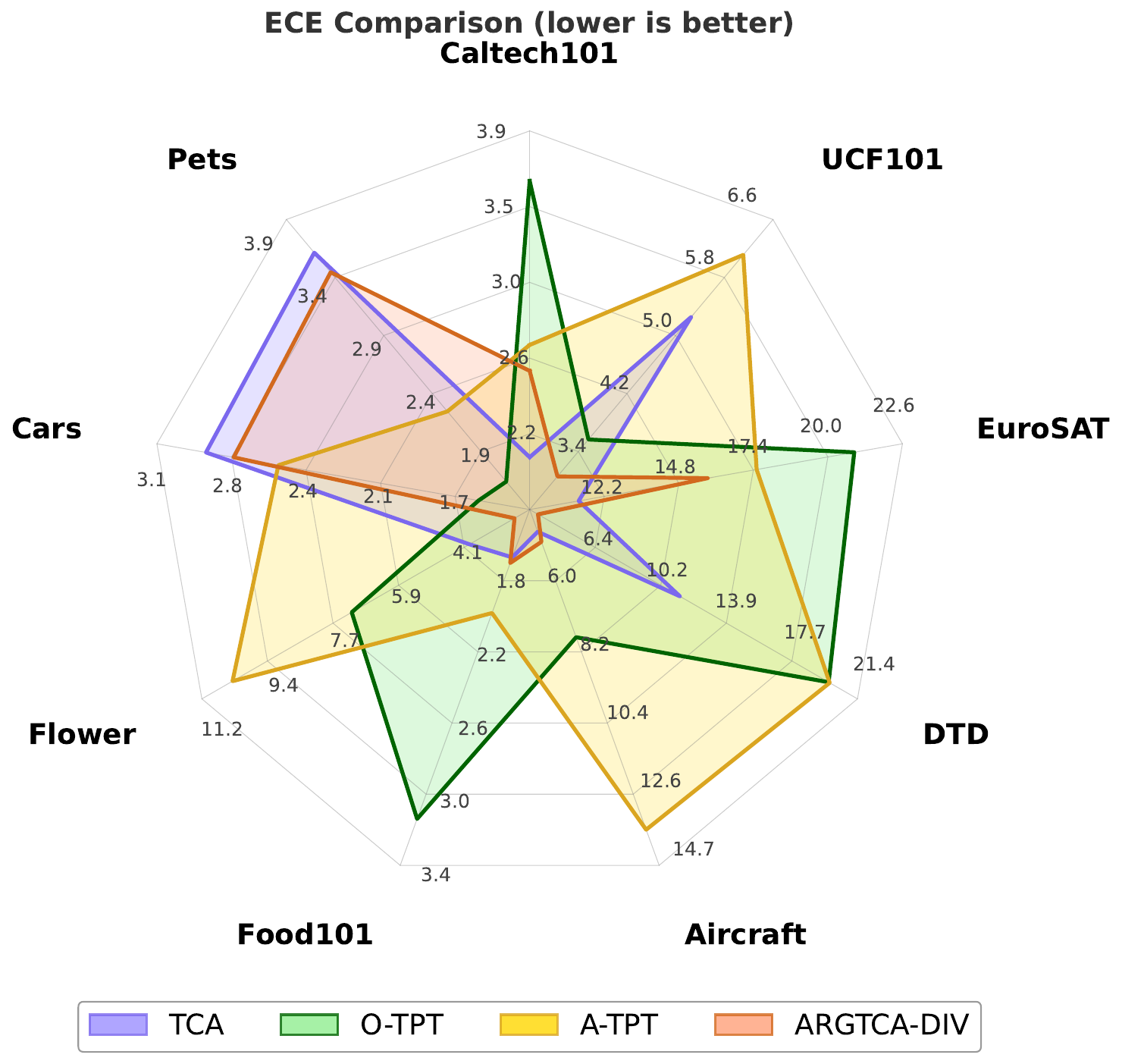}
    \caption{Radar plot comparing \ECE{} (lower is better, smaller area is better) of \TCA{}, \OTPT{}, A-TPT, and \ARGTCADiv{} (Ours) across nine datasets (ViT-B/16). \ARGTCADiv{} consistently occupies the innermost region, with the largest gains on Aircraft, Flower, and EuroSAT.}
    \label{fig:radar}
\end{figure}

Table~\ref{tab:results} 
shows that \ARGTCADiv{} achieves the best average \ECE{} of $4.45\%$, reducing miscalibration by $26.7\%$, $31.5\%$, $42.4\%$, and $48.0\%$ over \CTPT{}, \TCA{}, \OTPT{}, and A-TPT respectively, while improving average accuracy simultaneously. \ARGTCADisc{} achieves the highest average accuracy of $64.68\%$, with an \ECE{} of $5.90\%$ --- a $2.8\%$, $9.2\%$, $23.7\%$, and $31.0\%$ reduction over \CTPT{}, \TCA{}, \OTPT{}, and A-TPT respectively. Notably, A-TPT achieves competitive accuracy ($64.84\%$) but at a significantly higher calibration cost (\ECE{}: $8.55\%$).
\ARGTCADisc{} breaks this trade-off --- matching A-TPT's accuracy within $0.2\%$ while reducing its \ECE{} by $31\%$, 
confirming that graph-structured attribute reasoning provides consistent calibration gains.

\begin{table*}[h]
\scalebox{0.88}{
\begin{tabular}{llcccccccccc}
\toprule
{\textbf{Method}} & \rotatebox{90}{Metric}& \rotatebox{90}{Caltech} & \rotatebox{90}{Pets} & \rotatebox{90}{Cars} & \rotatebox{90}{Flower} & \rotatebox{90}{Food101} & \rotatebox{90}{Aircraft} & \rotatebox{90}{DTD} & \rotatebox{90}{EuroSAT} & \rotatebox{90}{UCF101} & \rotatebox{90}{\textbf{Average}} \\
\midrule\midrule
\multicolumn{12}{l}{\textbf{Pre-trained Backbone: CLIP Vit-B/16} | \textit{Embedding dimension: 1024-d}} \\
\midrule
TPT (NeurIPS'22)
& Acc. & 93.83  & 87.08  & 66.32  & 69.31  & 84.70  & 23.61  & 46.70  & 42.79  & 67.27  & 64.62 \\
& ECE  & 4.59  & 5.77  & 5.25  & 13.27 & 4.05  & 16.48  & 21.35  & 21.49  & 13.01  & 11.70 \\
\hdashline
C-TPT (ICLR'24)
& Acc. & 93.35  & 84.14  & 65.45  & 69.79  & 83.28  & 23.85  & 46.04  & 27.8  & 59.7  & 61.49 \\
& ECE  & 4.24  & 2.77  & 1.94  & 5.21 & 3.77  & \second{4.38}  & 12.27  & 15.16  & 3.89  & 6.07 \\
\hdashline
\TCA{} (ECML'25)     
& Acc. & 93.82 & 90.51 & 65.92 & 69.18 & 69.18 & 24.96 & 44.73 & 45.52 & 66.9  & 63.31\\
& ECE  & 2.56  & 6.3   & 7.85  & 3.67  & 5.28  & 4.52   & 11.26  & \best{11.35} & 5.25  & 6.50\\
\hdashline

O-TPT (CVPR'25)
& Acc.& 93.67& 88.55& 65.65& 69.96& 83.52& 23.46& 46.87& 42.85
& 66.90 & 64.60\\
& ECE& 3.62& \best{1.92}& \best{1.80}& 7.15& 3.13& 7.75& 19.75& 20.92 & 3.56&7.73\\
\hdashline
A-TPT (ICLR'26)& Acc. & 93.75 & 88.49 & 66.44 & 69.83 & 83.57 & 23.34 & 47.22 & 43.29 & 67.64 & 64.84\\
& ECE  & 2.67  & \second{2.23} & \second{2.54} & 10.38 & 2.20 & 13.64 & 19.81 & 17.53 & 6.11 &8.55\\
\hdashline

\ARGTCADiv{} (Ours) & Acc. 
& 92.82& 89.48& 65.07& 70.36& 83.64& 24.69& 47.34& 32.84& 69.63&63.99\\
& ECE  &\second{2.52}& 3.44& 2.75& \best{2.74}& \best{1.75}& 4.83& \best{3.17}& \second{15.83}& \best{3.05} &\best{4.45}\\
\hdashline

\ARGTCADisc{} (ours)
& Acc. & 93.19& 88.91& 66.47& 69.59& 83.85& 25.05& 46.63& 39.53& 68.70 &64.68\\
& ECE  & \best{1.62}& 3.37& 4.29& \second{3.57}& \second{2.06}& \best{4.25}& \second{6.36}& 19.81& \second{3.81} &\second{5.90}\\

\bottomrule
\end{tabular}}
\caption{Top-1 accuracy (Acc.\%, $\uparrow$) and Expected Calibration Error (\ECE{}, \%, $\downarrow$) on nine benchmarks on ViT-B/16. \best{red}: best per dataset and metric; \second{orange}: second best. }
\label{tab:results}
\end{table*}

\subsection{Calibration Quality: Confidence Distributions and Reliability}

\begin{figure}[h]
    \centering
    \begin{subfigure}[b]{0.48\linewidth}
        \includegraphics[width=\textwidth]{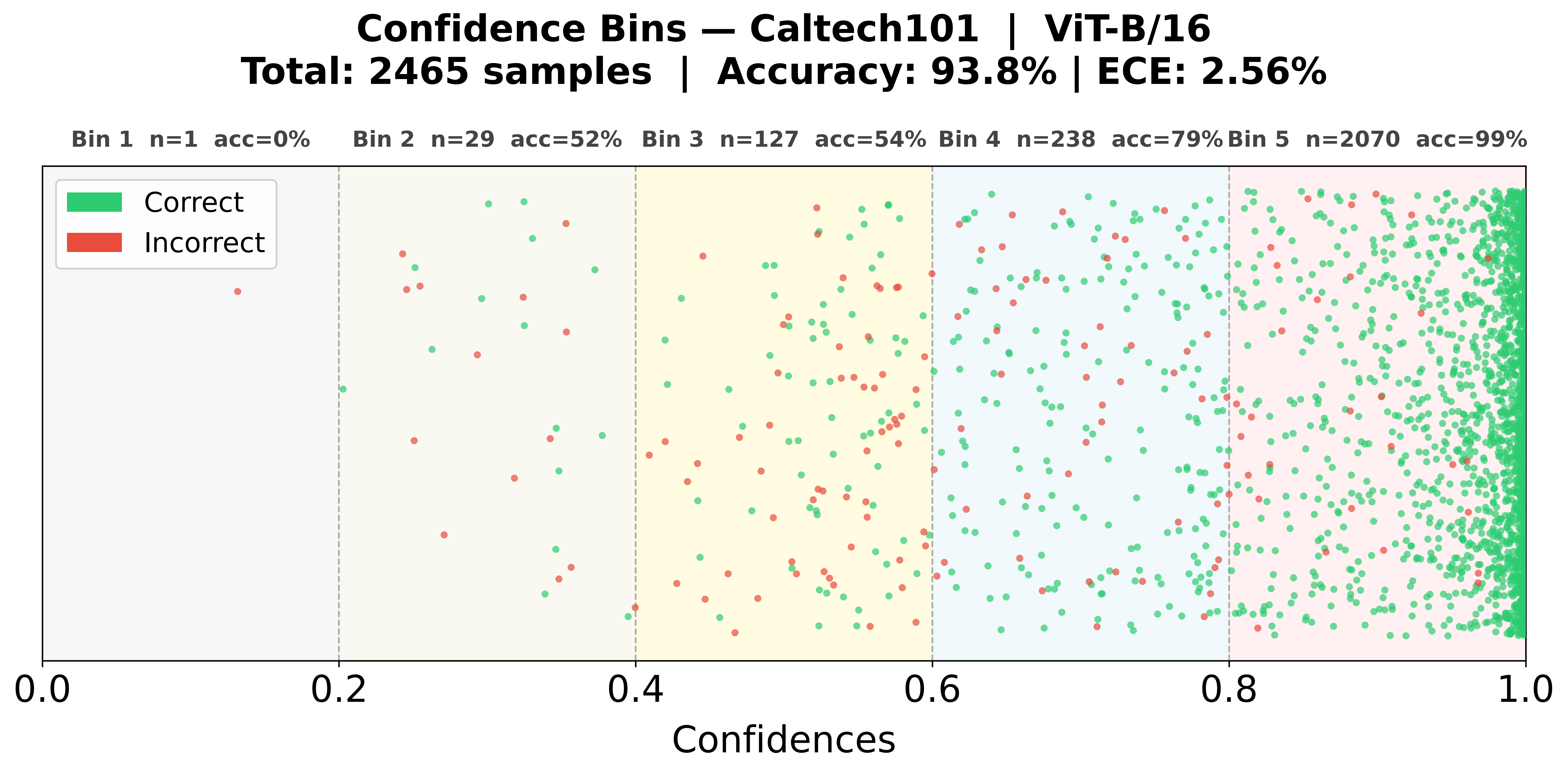}
        \caption{\TCA{}}
        \label{fig:conf_cal_tpt}
    \end{subfigure}
    \hfill
    \begin{subfigure}[b]{0.48\linewidth}
        \includegraphics[width=\textwidth]{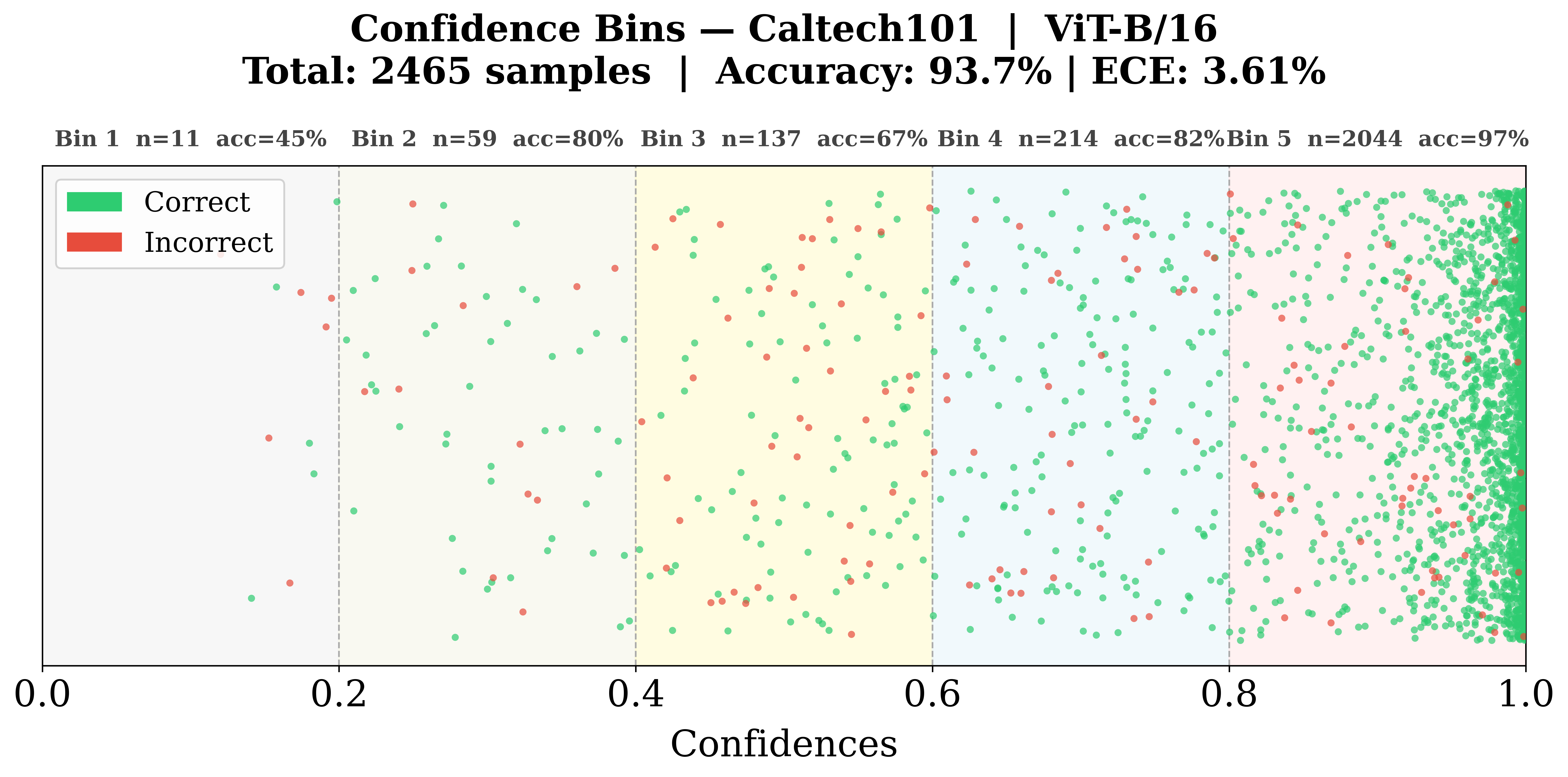}
        \caption{\OTPT{}}
        \label{fig:conf_cal_tca}
    \end{subfigure}
    \vspace{0.3em}
    \begin{subfigure}[b]{0.48\linewidth}
        \includegraphics[width=\textwidth]{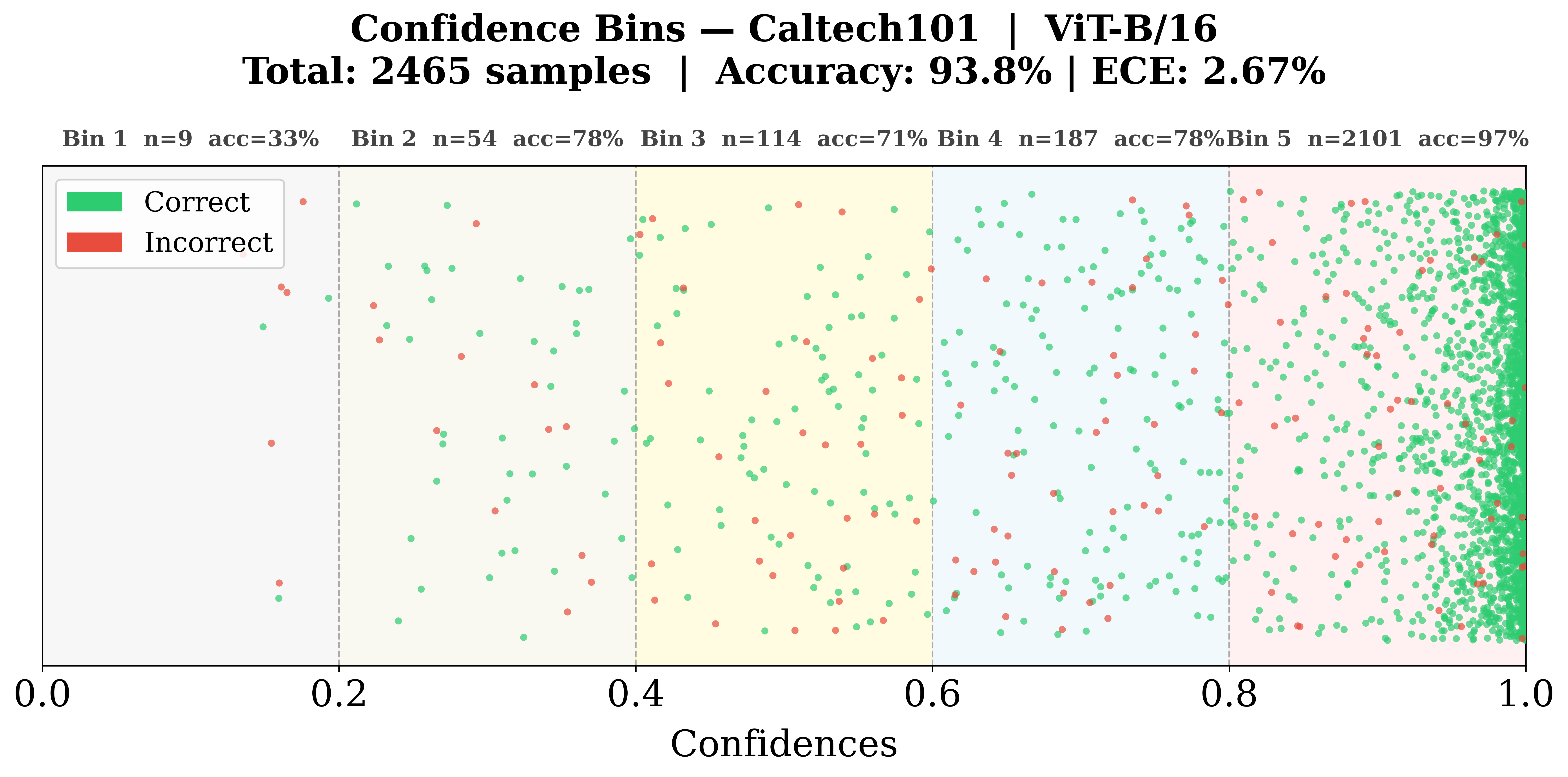}
        \caption{A-TPT}
        \label{fig:conf_cal_ctpt}
    \end{subfigure}
    \hfill
    \begin{subfigure}[b]{0.48\linewidth}
        \includegraphics[width=\textwidth]{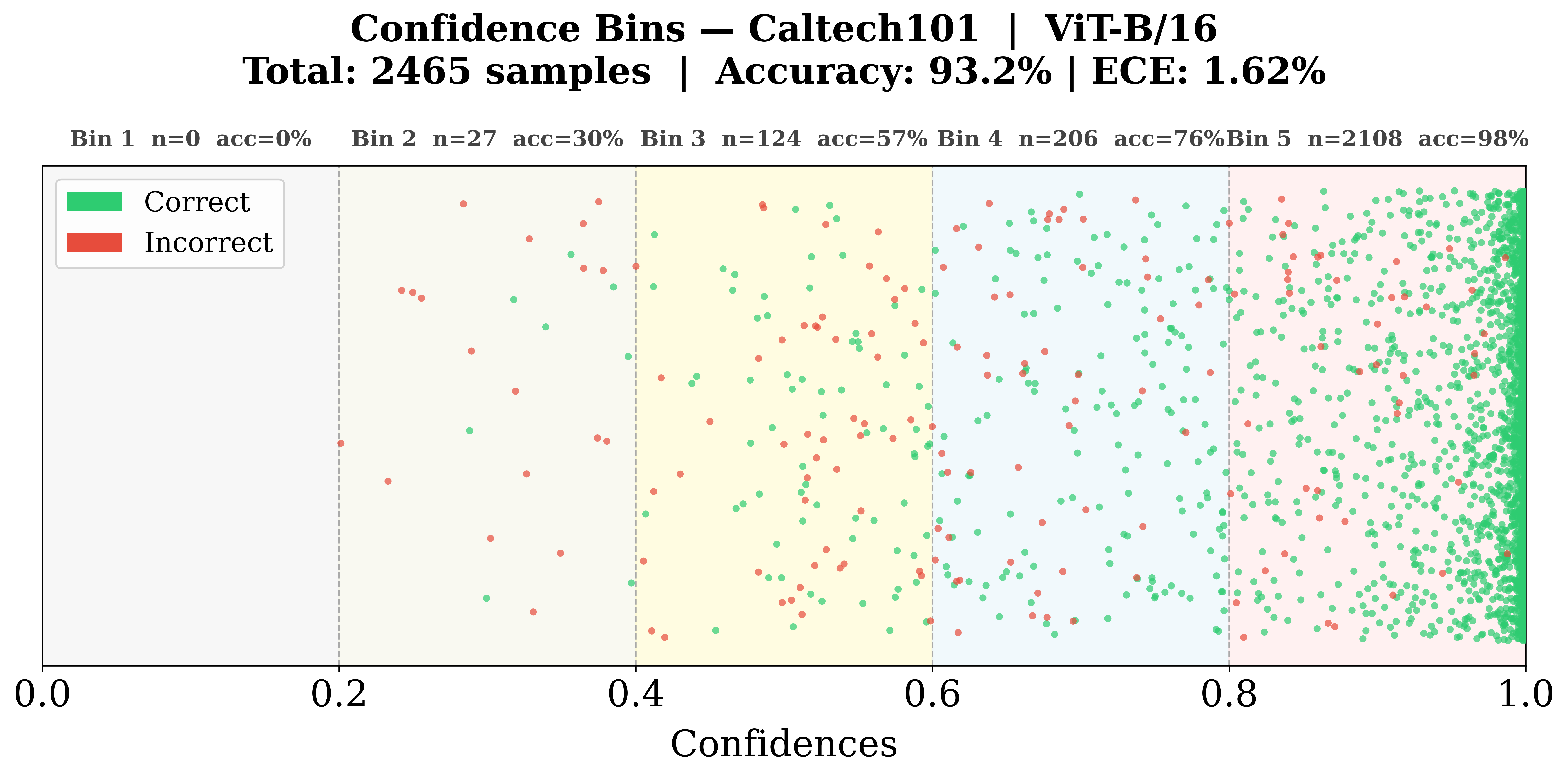}
        \caption{\ARGTCA{} (Ours)}
        \label{fig:conf_cal_argtca}
    \end{subfigure}
    \caption{Confidence bin plots on Caltech101 (ViT-B/16). Green/red
    indicate correct/incorrect predictions. \ARGTCADisc{} achieves the
    lowest ECE with the cleanest separation between correct and incorrect
    predictions across confidence bins.}
    \label{fig:conf_Caltech101}
\end{figure}

The confidence bin plots (Figure~\ref{fig:conf_Caltech101}) reveal that
\TCA{}, \OTPT{}, and A-TPT accumulate incorrect (red) predictions at high
confidence levels, indicating systematic overconfidence. \ARGTCADisc{}
achieves the lowest ECE of $1.62\%$ at $93.2\%$ accuracy, with incorrect
predictions concentrated at low confidence and correct predictions (green)
dominating the high-confidence region. By contrast, \OTPT{} (ECE: $3.61\%$)
and \TCA{} (ECE: $2.56\%$) exhibit red-green mixing in mid-confidence bins
(2--4), while A-TPT (ECE: $2.67\%$) partially improves but still
misclassifies samples at low confidence.

The radar plot (Figure~\ref{fig:radar}) confirms this trend across all nine
benchmarks: \ARGTCA{} consistently occupies the innermost region,
with gains most pronounced on Aircraft
(\ECE{}: $1.21\%$ vs.\ $8.19\%$ for A-TPT), Food101 ($1.75\%$
vs.\ $2.02\%$ for A-TPT), and Flower ($2.88\%$ vs.\ $10.38\%$ for
A-TPT). Reliability diagrams
(Figures~\ref{fig:reliability-caltech},~\ref{fig:reliability-all}) further
corroborate this, with \ARGTCA{} showing the closest alignment to the
calibration diagonal --- particularly in high-confidence bins
(0.8--1.0) --- confirming that graph-based attribute selection suppresses
overconfidence without inducing underconfidence.

\subsection{Accuracy--Calibration Trade-off}
\label{sec:results:tradeoff}
The two selection strategies encode complementary objectives.
\ARGTCADiv{} selects the most semantically complementary attribute pair
per class, maximizing angular spread in $\hL$ space. The resulting
diverse prompt initialization provides strong gradient signals for both
$\mathcal{L}_{\mathrm{intra}}$ and $\mathcal{L}_{\mathrm{inter}}$,
driving higher accuracy alongside well-calibrated text distributions.
\ARGTCADisc{} instead targets attributes maximally distant from
other-class embeddings, initializing prompts with high inter-class
separation and more conservative updates --- effective on datasets with
severe inter-class confusion (e.g., Caltech101, Aircraft) but less
consistent across all nine benchmarks.

\noindent\textbf{Recommendation:} \ARGTCADiv{} for broad cross-dataset
calibration; \ARGTCADisc{} when inter-class confusion dominates.

\begin{figure}[t]
    \centering
    \includegraphics[width=1\linewidth]{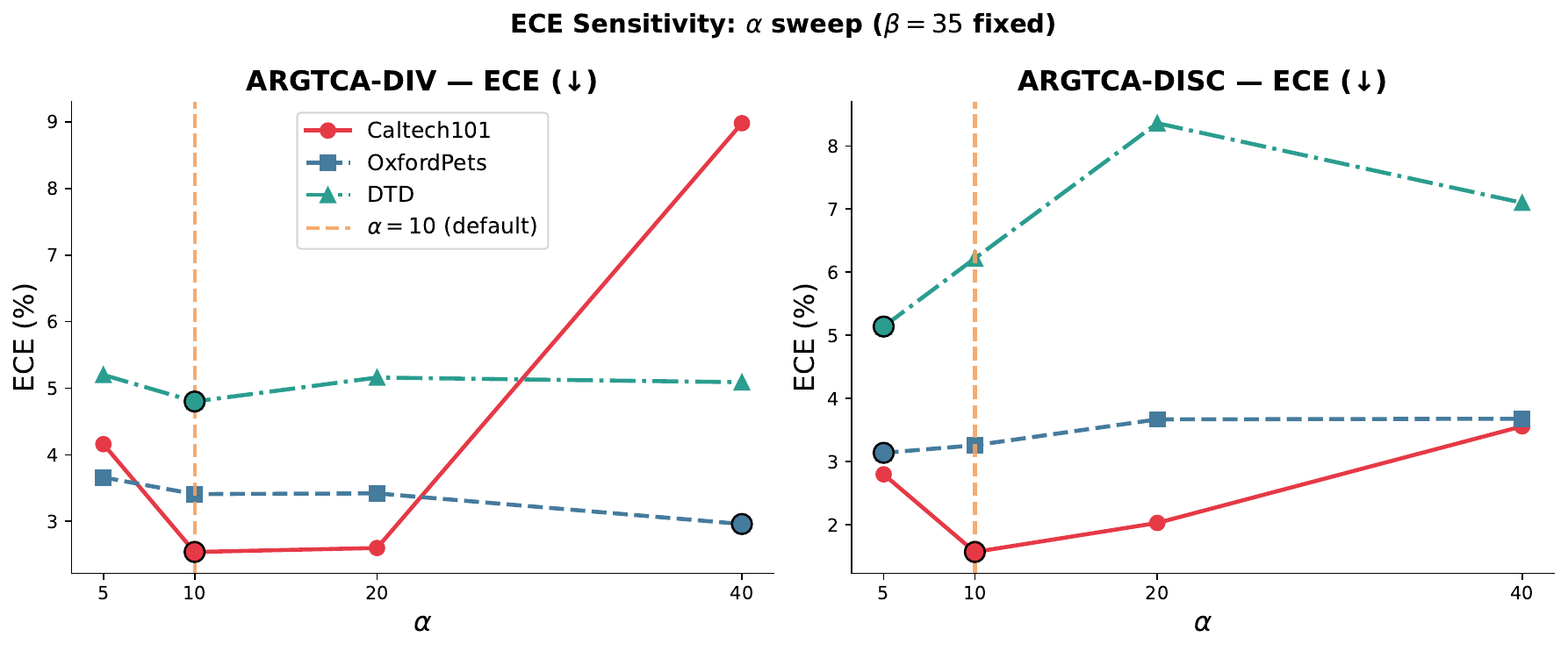}
    \caption{ECE vs.\ $\alpha$ (inter-class regularization
    weight) on Caltech101, OxfordPets, and DTD ($\beta{=}35$ fixed).}
    \label{fig:alpha}
\end{figure}

\begin{figure}[t]
    \centering
    \includegraphics[width=1\linewidth]{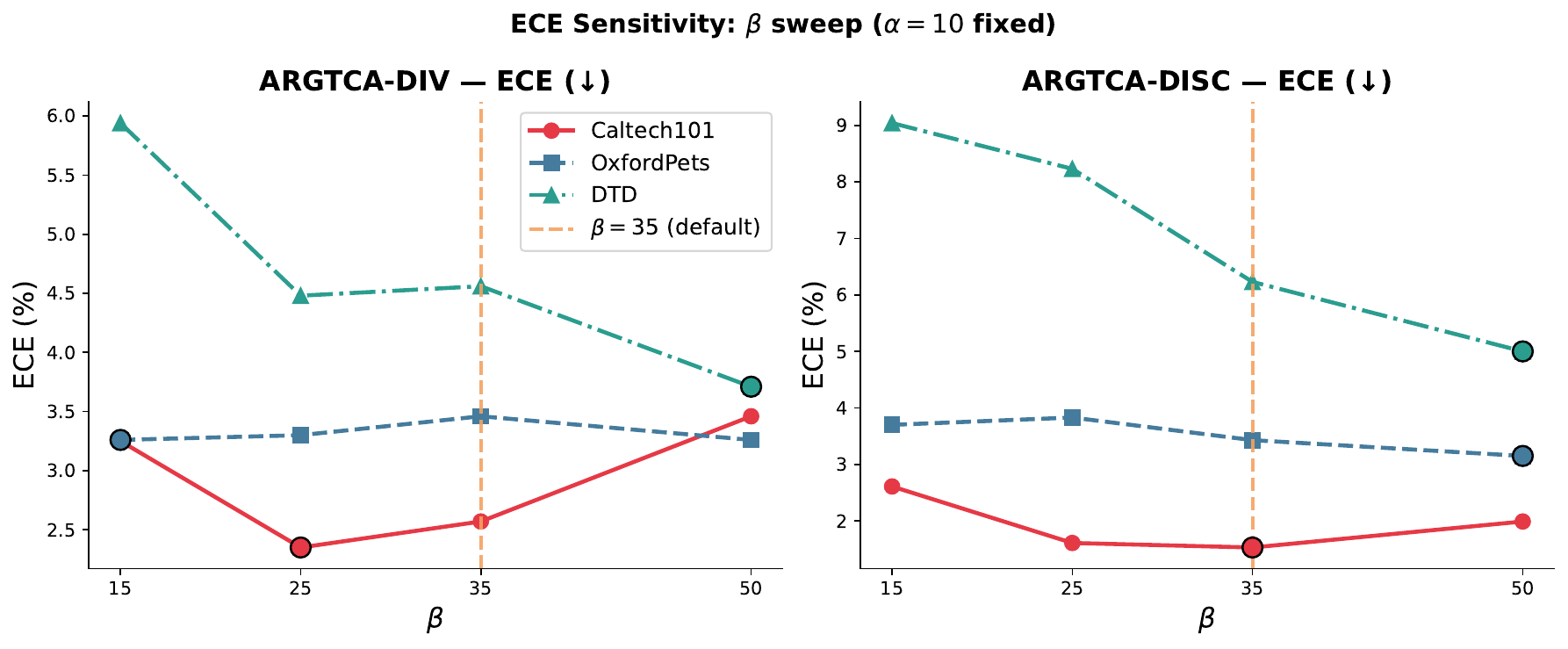}
    \caption{ECE vs.\ $\beta$ (intra-class regularization
    weight) on Caltech101, OxfordPets, and DTD ($\alpha{=}10$ fixed).}
    \label{fig:beta}
\end{figure}
\subsection{Ablation Studies}
\label{sec:results:ablation}
\begin{table}[t]
\centering
\small
\setlength{\tabcolsep}{4pt}
\scalebox{0.85}{
\begin{tabular}{ll cc cc cc}
\toprule
\multirow{2}{*}{Variant} & \multirow{2}{*}{Sel.}
  & \multicolumn{2}{c}{Caltech101}
  & \multicolumn{2}{c}{OxfordPets}
  & \multicolumn{2}{c}{DTD} \\
\cmidrule(lr){3-4}\cmidrule(lr){5-6}\cmidrule(lr){7-8}
& & Acc & ECE & Acc & ECE & Acc & ECE \\
\midrule

\multirow{2}{*}{\ARGTCA{} (Full)}
  & Div  & 92.82 & 2.52 & 89.48 & \second{3.44} & 47.34 & \best{3.17} \\
  & Disc & 93.19 & \best{1.62} & 88.91 & \best{3.37} & 46.63 & 6.36 \\
\hdashline

\multirow{2}{*}{(i) Zero edges}
  & Div  & 93.18 & 2.36 & 89.32 & 3.84 & 45.04 & \second{4.91} \\
  & Disc & 93.51 & \second{1.74} & 90.00 & 3.50 & 45.74 & 5.31 \\
\hdashline

\multirow{2}{*}{(ii) Raw EOS}
  & Div  & 92.21 & 2.73 & 88.74 & 3.87 & 46.81 & 5.65 \\
  & Disc & 93.39 & 2.60 & 88.99 & 3.71 & 45.98 & 8.95 \\
\hdashline

\multirow{2}{*}{(iii) Random sel.}
  & Div  & 93.10 & 2.53 & 88.63 & 3.44 & 45.92 & 5.16 \\
  & Disc & 92.25 & 2.43 & 88.50 & 3.55 & 44.27 & 6.33 \\
\bottomrule
\end{tabular}}
\caption{Ablation study on Caltech101, OxfordPets, and DTD (ViT-B/16). Acc\,(\%,\,$\uparrow$) / ECE\,(\%,\,$\downarrow$). \best{red}: best per column; \second{orange}: second best. Div\,=\,\ARGTCADiv{}; Disc\,=\,\ARGTCADisc{}.}
\label{tab:ablation}
\end{table}
All ablations use ViT-B/16 on Caltech101, OxfordPets, and DTD. We ablate three components of \ARGTCA{}: (i)~\textit{edge structure} ---
\SAG{} trained with $\mathcal{E}{=}\emptyset$, removing message passing
while retaining contrastive training and Div/Disc selection;
(ii)~\textit{\GAT{} training} --- Div/Disc applied directly to raw \CLIP{}
EOS embeddings $\mathbf{h}^{(0)}$, skipping Phase~1 entirely; and
(iii)~\textit{selection criterion} --- Phase~1 run in full but with random
attribute selection ($M'{=}2$, seed~42). We additionally sweep
$\alpha \in \{5,10,20,40\}$ ($\beta{=}35$ fixed) and
$\beta \in \{15,25,35,50\}$ ($\alpha{=}10$ fixed).
A detailed per-ablation description and hyperparameter sweep figures are in shown Appendix~\ref{app:ablation}.

The ablation results (Table~\ref{tab:ablation}) confirm the contribution
of each component. Removing edge structure~(i) degrades \ECE{} on DTD
for both Div ($3.17{\to}4.91$) and Disc ($6.36{\to}4.91$). Skipping \GAT{} training entirely~(ii) causes the largest degradation:
DTD \ECE{} rises to $5.65$ (Div) and $8.95$ (Disc), confirming that
relational contrastive training is critical for texture-heavy datasets.
Random selection~(iii) degrades DTD Div \ECE{} from $3.17$ to $5.16$
with minimal accuracy impact, directly isolating the geometric Div/Disc
criteria as the operative calibration mechanism.

\begin{table}[h]
\centering
\small
\setlength{\tabcolsep}{4pt}
\renewcommand{\arraystretch}{1.2}
\scalebox{0.67}{
\begin{tabular}{@{}ll p{2.5cm} p{2.5cm} p{2.5cm}@{}}
\toprule
Dataset & Class & \TCA{} (top-2) & \ARGTCADiv{} & \ARGTCADisc{} \\
\midrule
\multicolumn{5}{l}{\textit{Caltech101}} \\
& airplane     & aircraft, travel          & transportation, altitude & jet, airline \\
& camera       & image, digital            & digital, vintage         & dslr, digital \\
& chair        & design, chair             & design, director's       & elegant, chair \\
& face         & images, human             & natural, adults          & emotion, casual \\
& motorbike    & motorbike, pillion        & seat, suspension         & suspension, motorbike \\
\midrule
\multicolumn{5}{l}{\textit{OxfordPets}} \\
& Abyssinian   & Abyssinian, cats          & restful, observant       & cats, feline \\
& Boxer        & Dog, Boxer                & Athletic, Friendly       & Boxer, Pet \\
& Maine Coon   & Feline, Cat               & Tabby, Fluffy            & Kitten, Coon \\
& Persian      & Adorable, Persian         & Adorable, Lying          & Persian, Adorable \\
& Shiba Inu    & Shiba, Dog                & Cute, Foxy               & Inu, Shiba \\
\midrule
\multicolumn{5}{l}{\textit{DTD}} \\
& banded       & design, casual            & material, colorful       & casual, colorful \\
& bumpy        & uneven, unequal           & unequal, geometric       & knobby, indented \\
& crystalline  & natural, crystals         & natural, glittering      & shimmering, crystals \\
& dotted       & variation, design         & variation, fashion       & dots, stylized \\
& woven        & interwoven, craftsmanship & material, organic        & handwoven, craftsmanship \\
\bottomrule
\end{tabular}}
\caption{Representative attribute selection per class across three datasets.
(\ARGTCADiv{}, \ARGTCADisc{}) avoid near-synonymous
descriptors that cluster tightly in \CLIP{} embedding space, unlike \TCA{}.}
\label{tab:attr_comparison}
\end{table}

\paragraph{Attributes selected by TCA vs.\ \ARGTCA{}.}
\TCA{} selects top-$M'$ attributes by cosine similarity to the class name, surfacing near-synonymous descriptors that cluster tightly on the hypersphere. \ARGTCADiv{} selects the maximally complementary pair in $\hL$ space; \ARGTCADisc{} selects the pair most distant from all other-class embeddings. Per-class examples are in Table~\ref{tab:attr_comparison}.

\paragraph{Hyperparameter sensitivity ($\alpha$, $\beta$).}
Both weights are stable across $\alpha \in \{5,10,20,40\}$ and $\beta \in \{15,25,35,50\}$: ECE varies by at most ${\sim}2\%$ across the full sweep, with the ViT-B/16 defaults ($\alpha{=}10$, $\beta{=}35$) lying in the stable plateau (see Figures~\ref{fig:alpha}--\ref{fig:beta}). Full ECE sweep for both selection strategies are provided in Appendix~\ref{app:ablation}.

\FloatBarrier
\section{Conclusion}
\label{sec:conclusion}

We present \ARGTCA{}, a test-time calibration framework for \CLIP{} that
replaces flat attribute selection in \TCA{} with graph-relational
reasoning. A \GAT{} trained via supervised contrastive loss over the
Symbolic Attribute Graph (\SAG{}) produces attribute embeddings encoding
intra-class diversity and inter-class discriminability, giving rise to
two selection criteria: \ARGTCADiv{} and \ARGTCADisc{}.
Across nine benchmarks, \ARGTCADiv{} achieves
the lowest average \ECE{} ($4.45$ ), reducing
miscalibration by up to $48\%$. Ablations confirm that edge topology, contrastive training, and geometric selection each contribute independently.
Future works include a unified selection objective jointly optimiing diversity and discriminability, and cross-dataset graph transfer to remove the per-dataset offline build requirement.

\section*{Limitations}


\paragraph{LLM attribute quality.}
The method inherits \TCA{}'s dependence on \LLM{}-generated attribute quality.
Domains where \LLM{}s produce generic or noisy attributes (e.g., satellite
imagery) remain challenging.  Richer, domain-specific attribute generation
strategies may be needed for such cases.

\paragraph{Accuracy--calibration tradeoff.}
\ARGTCADiv{} and \ARGTCADisc{} are complementary but not jointly optimal:
across benchmarks, \ARGTCADiv{} tends to improve accuracy while \ARGTCADisc{}
tends to reduce \ECE{}, and no single variant dominates on both metrics
simultaneously.  Practitioners must choose a selection strategy based on
deployment priority, which limits plug-and-play usability without prior
knowledge of the target domain.

\paragraph{Dataset-specific graph.}
The \SAG{} and \GAT{} are trained independently per dataset, so applying
\ARGTCA{} to a new benchmark requires re-running Phase~1.  The learned
graph structure does not transfer across datasets or domains, and Phase~1
cannot be run when the class vocabulary is unavailable at build time,
constraining applicability in fully zero-shot settings.

\newpage


\bibliography{custom}

@inproceedings{radford2021learning,
  author    = {Radford, Alec and Kim, Jong Wook and Hallacy, Chris and
               Ramesh, Aditya and Goh, Gabriel and Agarwal, Sandhini and
               Sastry, Girish and Askell, Amanda and Mishkin, Pamela and
               Clark, Jack and Krueger, Gretchen and Sutskever, Ilya},
  title     = {Learning Transferable Visual Models From Natural Language Supervision},
  booktitle = {Proceedings of the 38th International Conference on Machine Learning},
  series    = {ICML},
  pages     = {8748--8763},
  year      = {2021},
  publisher = {PMLR},
}

@inproceedings{jia2021scaling,
  author    = {Jia, Chao and Yang, Yinfei and Xia, Ye and Chen, Yi-Ting and
               Parekh, Zarana and Pham, Hieu and Le, Quoc V. and Sung, Yunhsuan and
               Li, Zhen and Duerig, Tom},
  title     = {Scaling Up Visual and Vision-Language Representation Learning with Noisy Text Supervision},
  booktitle = {Proceedings of the 38th International Conference on Machine Learning},
  series    = {ICML},
  year      = {2021},
}

@article{zhou2022learning,
  author    = {Zhou, Kaiyang and Yang, Jingkang and Loy, Chen Change and Liu, Ziwei},
  title     = {Learning to Prompt for Vision-Language Models},
  journal   = {International Journal of Computer Vision},
  volume    = {130},
  pages     = {2337--2348},
  year      = {2022},
}

@inproceedings{zhou2022conditional,
  author    = {Zhou, Kaiyang and Yang, Jingkang and Loy, Chen Change and Liu, Ziwei},
  title     = {Conditional Prompt Learning for Vision-Language Models},
  booktitle = {Proceedings of the IEEE/CVF Conference on Computer Vision and
               Pattern Recognition},
  series    = {CVPR},
  pages     = {16816--16825},
  year      = {2022},
}

@inproceedings{shu2022test,
  author    = {Shu, Manli and Nie, Weili and Huang, De-An and Yu, Zhiding and
               Goldstein, Tom and Anandkumar, Anima and Xiao, Chaowei},
  title     = {Test-Time Prompt Tuning for Zero-Shot Generalization in
               Vision-Language Models},
  booktitle = {Advances in Neural Information Processing Systems},
  series    = {NeurIPS},
  volume    = {35},
  pages     = {14274--14289},
  year      = {2022},
}

@inproceedings{feng2023diverse,
  author    = {Feng, Chun-Mei and Yu, Kai and Liu, Yong and Khan, Salman and
               Zhong, Wangmeng},
  title     = {Diverse Data Augmentation with Diffusions for Effective
               Test-Time Prompt Tuning},
  booktitle = {Proceedings of the IEEE/CVF International Conference on Computer
               Vision},
  series    = {ICCV},
  pages     = {2704--2714},
  year      = {2023},
}

@inproceedings{mirza2024promptalign,
  author    = {Mirza, Muhammad Uzair and Khalid, Fahad Shahbaz and
               Siddiqui, Fawad and Khan, Salman},
  title     = {LAFTER: Label-Free Test-Time Adaptation with Language Feedback
               for Robust Visual Prompt Tuning},
  booktitle = {Advances in Neural Information Processing Systems},
  series    = {NeurIPS},
  year      = {2023},
}

@inproceedings{guo2017calibration,
  author    = {Guo, Chuan and Pleiss, Geoff and Sun, Yu and Weinberger, Kilian Q.},
  title     = {On Calibration of Modern Neural Networks},
  booktitle = {Proceedings of the 34th International Conference on Machine Learning},
  series    = {ICML},
  pages     = {1321--1330},
  year      = {2017},
}

@inproceedings{yoon2024ctpt,
  author    = {Yoon, Jihoon and Oh, Jae-hun and Kim, Byungho and
               Kweon, In So and Choi, Jongwoo},
  title     = {{C-TPT}: Calibrated Test-Time Prompt Tuning for Vision-Language
               Models via Text Feature Dispersion},
  booktitle = {International Conference on Learning Representations},
  series    = {ICLR},
  year      = {2024},
}

@article{hebbalaguppe2025tca,
  author    = {Hebbalaguppe, Ramya and Kandar, Tamoghno and Nagpal, Abhinav and
               Arora, Chetan},
  title     = {Prompting without Panic: Attribute-Aware, Zero-Shot,
               Test-Time Calibration},
  journal   = {arXiv preprint arXiv:2506.22819},
  year      = {2025},
}

@inproceedings{sharifdeen2025otpt,
  author    = {Sharifdeen, A. and others},
  title     = {{O-TPT}: Orthogonality-Regularized Test-Time Prompt Tuning for
               Zero-Shot Generalization in Vision-Language Models},
  booktitle = {International Conference on Learning Representations},
  series    = {ICLR},
  year      = {2025},
}

@inproceedings{platt1999probabilistic,
  author    = {Platt, John C.},
  title     = {Probabilistic Outputs for Support Vector Machines and Comparisons
               to Regularized Likelihood Methods},
  booktitle = {Advances in Large Margin Classifiers},
  pages     = {61--74},
  year      = {1999},
  publisher = {MIT Press},
}

@inproceedings{velickovic2018graph,
  author    = {Veli{\v{c}}kovi{\'c}, Petar and Cucurull, Guillem and
               Casanova, Arantxa and Romero, Adriana and Li{\`o}, Pietro and
               Bengio, Yoshua},
  title     = {Graph Attention Networks},
  booktitle = {International Conference on Learning Representations},
  series    = {ICLR},
  year      = {2018},
}

@inproceedings{kipf2017semi,
  author    = {Kipf, Thomas N. and Welling, Max},
  title     = {Semi-Supervised Classification with Graph Convolutional Networks},
  booktitle = {International Conference on Learning Representations},
  series    = {ICLR},
  year      = {2017},
}

@inproceedings{li2023graphadapter,
  author    = {Li, Xin and Lian, Wenxuan and Lu, Zhixin and Bao, Jingwen and
               Liu, Jiaming and Li, Junliang and Lan, Jiaxin and Zhu, Lingling},
  title     = {{GraphAdapter}: Tuning Vision-Language Models With Dual Knowledge Graph},
  booktitle = {Advances in Neural Information Processing Systems},
  series    = {NeurIPS},
  year      = {2023},
}

@inproceedings{zheng2023hgclip,
  author    = {Zheng, Xiaobao and Ji, Liqiang and Hong, Changhao and Ying, Nong
               and Liu, Mengqi and Zhang, Zheng},
  title     = {{HGCLIP}: Exploring Vision-Language Models with Graph Representations
               for Hierarchical Understanding},
  booktitle = {Proceedings of the IEEE/CVF Conference on Computer Vision and
               Pattern Recognition},
  series    = {CVPR},
  year      = {2024},
}

@ARTICLE{zhou2024vcgprompt,
       author = {{Wang}, Mengjia and {Liu}, Fang and {Jiao}, Licheng and {Li}, Shuo and {Li}, Lingling and {Chen}, Puhua and {Liu}, Xu and {Ma}, Wenping},
        title = "{VCGPrompt: Visual Concept Graph-Aware Prompt Learning for Vision-Language Models}",
      journal = {Pattern Recognition},
         year = 2026,
        month = feb,
       volume = {170},
        pages = {112012},
}

@inproceedings{khosla2020supervised,
  author    = {Khosla, Prannay and Tian, Yonglong and Wang, Huiwen and
               Liu, Chen and Valmadre, Jack and Tian, Chen and
               Norouzi, Mohammad and others},
  title     = {Supervised Contrastive Learning},
  booktitle = {Advances in Neural Information Processing Systems},
  series    = {NeurIPS},
  volume    = {33},
  pages     = {18661--18673},
  year      = {2020},
}

@inproceedings{fei2004learning,
  author    = {Fei-Fei, Li and Fergus, Rob and Perona, Pietro},
  title     = {Learning Generative Visual Models from Few Training Examples},
  booktitle = {Proceedings of the CVPR 2004 Workshop on Generative-Model
               Based Vision},
  year      = {2004},
}

@inproceedings{cimpoi2014describing,
  author    = {Cimpoi, Mircea and Maji, Subhransu and Kokkinos, Iasonas and
               Mohamed, Sammy and Vedaldi, Andrea},
  title     = {Describing Textures in the Wild},
  booktitle = {Proceedings of the IEEE/CVF Conference on Computer Vision and
               Pattern Recognition},
  pages     = {3606--3613},
  year      = {2014},
}

@inproceedings{bossard2014food,
  author    = {Bossard, Lukas and Guillaumin, Matthieu and Van Gool, Luc},
  title     = {{Food-101} -- Mining Discriminative Components with Random Forests},
  booktitle = {Proceedings of the 13th European Conference on Computer Vision},
  series    = {ECCV},
  pages     = {446--461},
  year      = {2014},
}

@inproceedings{parkhi2012cats,
  author    = {Parkhi, Omkar M. and Vedaldi, Andrea and Zisserman, Andrew and
               Jawahar, C. V.},
  title     = {Cats and Dogs},
  booktitle = {Proceedings of the IEEE/CVF Conference on Computer Vision and
               Pattern Recognition},
  pages     = {3498--3505},
  year      = {2012},
}

@article{soomro2012ucf101,
  author    = {Soomro, Khurram and Zamir, Amir Roshan and Shah, Mubarak},
  title     = {{UCF101}: A Dataset of 101 Human Actions Classes from Videos
               in the Wild},
  journal   = {arXiv preprint arXiv:1212.0402},
  year      = {2012},
}

@article{helber2019eurosat,
  author    = {Helber, Patrick and Bischke, Benjamin and Dengel, Andreas and
               Borth, Damian},
  title     = {{EuroSAT}: A Novel Dataset and Deep Learning Benchmark for
               Land Use and Land Cover Classification},
  journal   = {IEEE Journal of Selected Topics in Applied Earth Observations
               and Remote Sensing},
  volume    = {12},
  number    = {7},
  pages     = {2217--2226},
  year      = {2019},
}

@inproceedings{he2016deep,
  author    = {He, Kaiming and Zhang, Xiangyu and Ren, Shaoqing and Sun, Jian},
  title     = {Deep Residual Learning for Image Recognition},
  booktitle = {Proceedings of the IEEE/CVF Conference on Computer Vision and
               Pattern Recognition},
  pages     = {770--778},
  year      = {2016},
}

@misc{atpt,
      title={A-TPT: Angular Diversity Calibration Properties for Test-Time Prompt Tuning of Vision-Language Models}, 
      author={Shihab Aaqil Ahamed and Udaya S. K. P. Miriya Thanthrige and Ranga Rodrigo and Muhammad Haris Khan},
      booktitle = {International Conference on Learning Representations},
  series    = {ICLR},
  year={2026},
}

@inproceedings{krause2013cars,
  title     = {3{D} Object Representations for Fine-Grained Categorization},
  author    = {Krause, Jonathan and Stark, Michael and Deng, Jia and Fei-Fei, Li},
  booktitle = {ICCV Workshop on 3D Representation and Recognition},
  year      = {2013}
}

@article{maji2013aircraft,
  author       = {Subhransu Maji and
                  Esa Rahtu and
                  Juho Kannala and
                  Matthew B. Blaschko and
                  Andrea Vedaldi},
  title        = {Fine-Grained Visual Classification of Aircraft},
  journal      = {CoRR},
  volume       = {abs/1306.5151},
  year         = {2013},
}

@inproceedings{nilsback2008flowers,
  title     = {Automated Flower Classification over a Large Number of Classes},
  author    = {Nilsback, Maria-Elena and Zisserman, Andrew},
  booktitle = {Indian Conference on Computer Vision, Graphics and Image Processing},
  year      = {2008}
}
\newpage
\begin{table*}[t]
\centering
\renewcommand{\arraystretch}{1.35}
\setlength{\tabcolsep}{6pt}
\caption*{\large\textbf{Appendix Contents}}
\begin{tabular}{@{}p{0.72\linewidth}r@{}}
\toprule
\textbf{Section} & \textbf{Page} \\
\midrule

\hyperref[datasets]{\textbf{A\; Datasets}}
  & \pageref{datasets} \\[4pt]

\hyperref[app:architecture]{\textbf{B\; GAT Architecture Details}}
  & \pageref{app:architecture} \\
\quad\hyperref[app:architecture]{\textit{Attention and Aggregation}}
  & \pageref{app:architecture} \\
\quad\hyperref[app:architecture]{\textit{Residual, Normalisation, and Projection}}
  & \pageref{app:architecture} \\
\quad\hyperref[app:architecture]{\textit{Node Attention Score}}
  & \pageref{app:architecture} \\[4pt]

\hyperref[app:ablation]{\textbf{C\; Ablation Studies}}
  & \pageref{app:ablation} \\
\quad\hyperref[app:ablation]{\textit{C.1\; Attribute Comparison: TCA vs.\ \ARGTCADiv{} vs.\ \ARGTCADisc{}}}
  & \pageref{app:ablation} \\
\quad\hyperref[app:ablation]{\textit{C.2\; Zero Graph Edges}}
  & \pageref{app:ablation} \\
\quad\hyperref[app:ablation]{\textit{C.3\; No GAT Training (Raw CLIP EOS Embeddings)}}
  & \pageref{app:ablation} \\
\quad\hyperref[app:ablation]{\textit{C.4\; Random Attribute Selection}}
  & \pageref{app:ablation} \\
\quad\hyperref[app:ablation]{\textit{C.5\; Hyperparameter Sensitivity ($\alpha$, $\beta$)}}
  & \pageref{app:ablation} \\[4pt]

\hyperref[tab:prompt_breakdown]{\textbf{D\; Prompt Token Layout}}
  & \pageref{tab:prompt_breakdown} \\[4pt]
  
\hyperref[tab:notations]{\textbf{E\; Notation Reference}}
  & \pageref{tab:notations} \\[4pt]
  
\midrule
\multicolumn{2}{@{}l}{\textbf{Figures}} \\[2pt]

\hyperref[fig:tca_pipeline]{TCA Pipeline Overview}
  & \pageref{fig:tca_pipeline} \\
\hyperref[fig:reliability-all]{Reliability Diagrams Across All Datasets (ViT-B/16)}
  & \pageref{fig:reliability-all} \\[4pt]

\midrule
\multicolumn{2}{@{}l}{\textbf{Tables}} \\[2pt]

\hyperref[tab:attr_comparison]{Attribute Comparison: TCA vs.\ \ARGTCADiv{} vs.\ \ARGTCADisc{}}
  & \pageref{tab:attr_comparison} \\
\hyperref[tab:ablation]{Ablation Results}
  & \pageref{tab:ablation} \\
  
\hyperref[tab:prompt_breakdown]{Prompt Token Layout — \ARGTCA{}}
  & \pageref{tab:prompt_breakdown} \\
\hyperref[tab:notations]{Notation Reference}
  & \pageref{tab:notations} \\

\bottomrule
\end{tabular}
\end{table*}
\clearpage
\appendix
\begin{table*}[t]
\begin{center}
\scalebox{0.9}{
\begin{tabular}{cll}
\toprule
Position & Content & Notes\\
\midrule
0 & SOS token embedding & Frozen\\
$1$--$n_{\text{ctx}}$ & Soft prompt $\mathbf{p} \in \R^{n_{\text{ctx}} \times D_t}$ & \textbf{Learnable (tuned at test time)}\\
$n_{\text{ctx}}+1$ & ${a}_m^k$ & \textbf{graph selected attribute}\\
$n_{\text{ctx}}+2$ -- $L\!-\!1$ & Class tokens + EOS + padding & Frozen\\
\bottomrule
\end{tabular}}
\end{center}
\caption{Full positional breakdown of a prompt in \ARGTCA{}.}
\label{tab:prompt_breakdown}
\end{table*}
\section{Datasets}
\label{datasets}
We evaluate on nine benchmarks: Caltech101~\cite{fei2004learning} (general objects), StanfordCars~\cite{krause2013cars}, FGVC-Aircraft~\cite{maji2013aircraft},
OxfordPets~\cite{parkhi2012cats}, and Flowers102~\cite{nilsback2008flowers}
(fine-grained recognition), Food101~\cite{bossard2014food} (food categories),
DTD~\cite{cimpoi2014describing} (perceptual textures), UCF101~\cite{soomro2012ucf101}
(action recognition), and EuroSAT~\cite{helber2019eurosat} (satellite land-use). We follow the standard \cite{zhou2022learning} test splits used in \TCA{}~\citep{hebbalaguppe2025tca}, evaluating on the test set only with no labeled data.
\begin{figure*}
    \centering
    \includegraphics[width=1\linewidth]{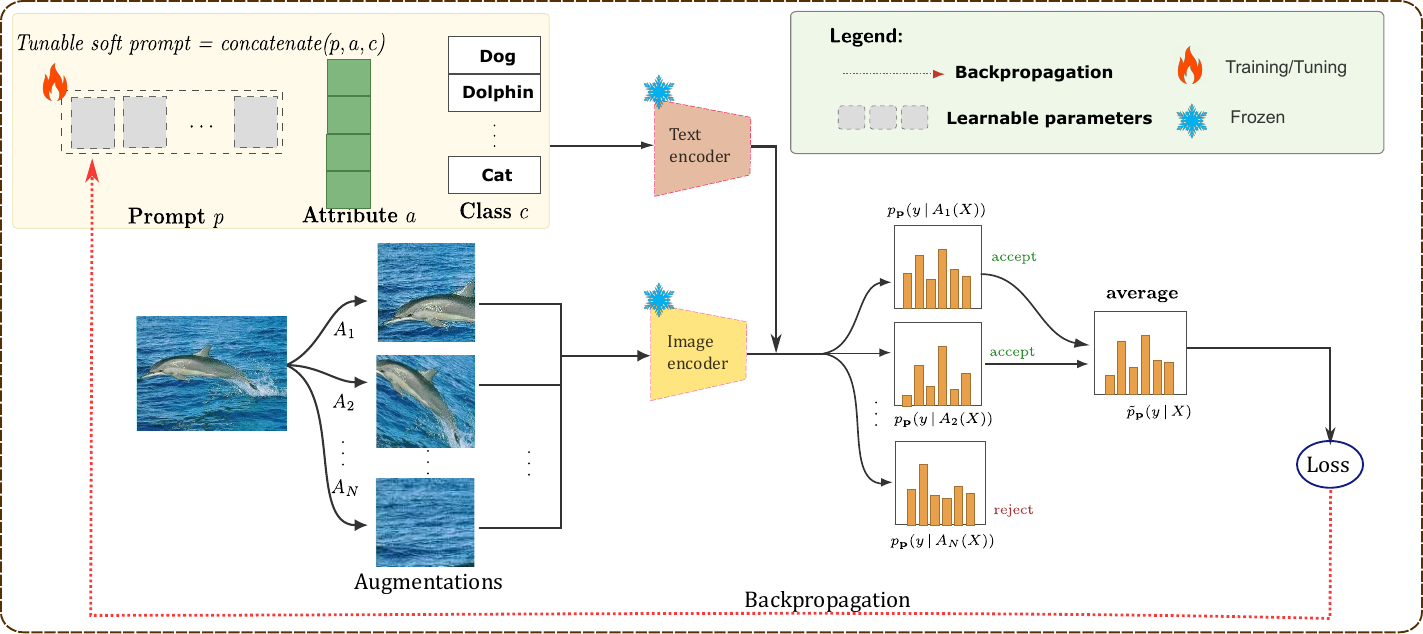}
    \caption{Test time prompt calibration}
    \label{fig:tca_pipeline}
\end{figure*}

\begin{figure*}[h]
\centering

\newcommand{\mw}{0.15\linewidth}
\begin{tabular}{cccccc}
\textbf{TPT} & \textbf{\TCA{}} & \textbf{\CTPT{}} & \textbf{A-TPT} & \textbf{O-TPT} & \textbf{\ARGTCA{} (Ours)} \\

\multicolumn{6}{l}{\small\textit{DTD}} \\[0.5pt]
\includegraphics[width=\mw]{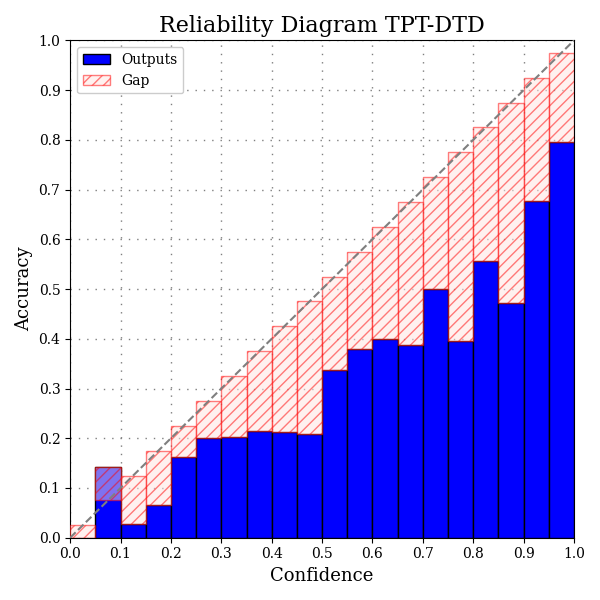} &
\includegraphics[width=\mw]{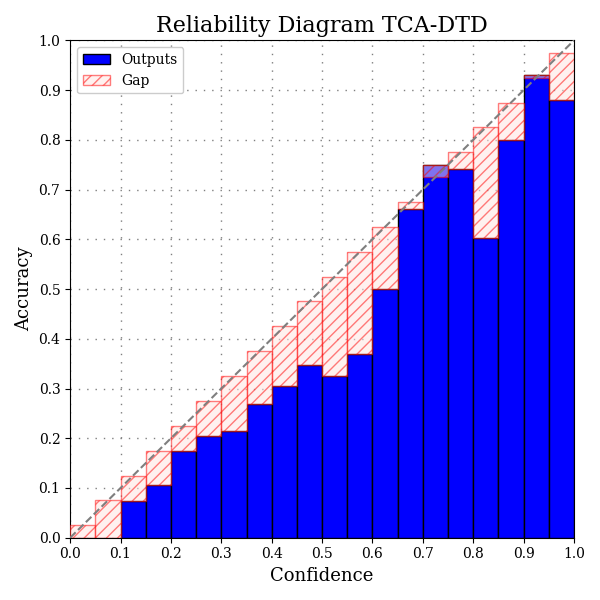} &
\includegraphics[width=\mw]{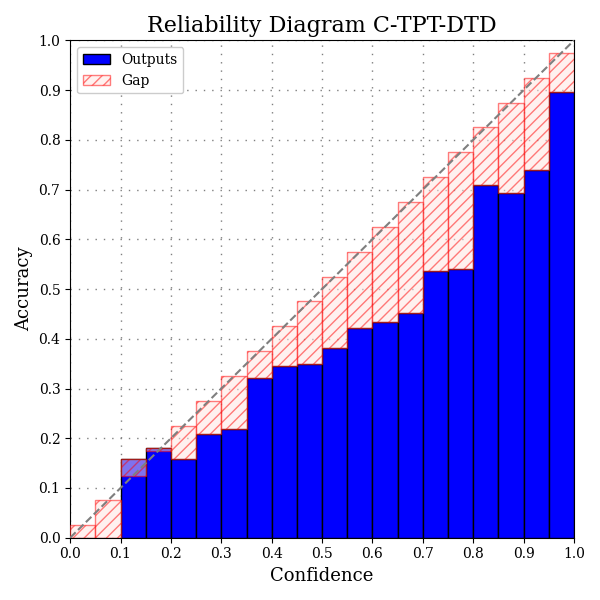} &
\includegraphics[width=\mw]{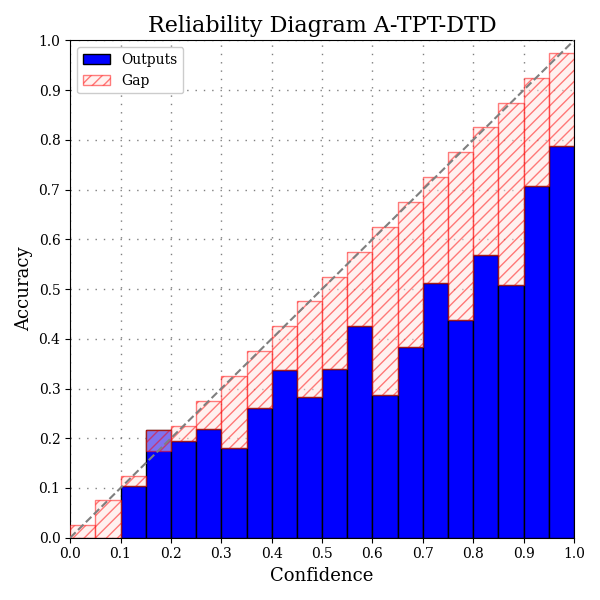} &
\includegraphics[width=\mw]{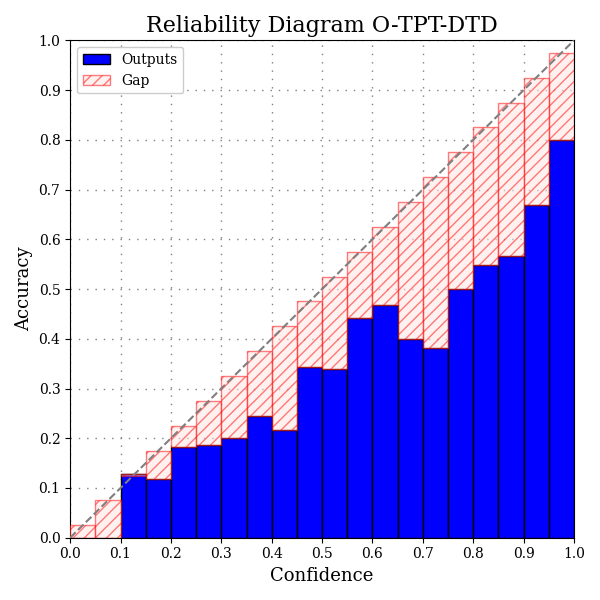} &
\includegraphics[width=\mw]{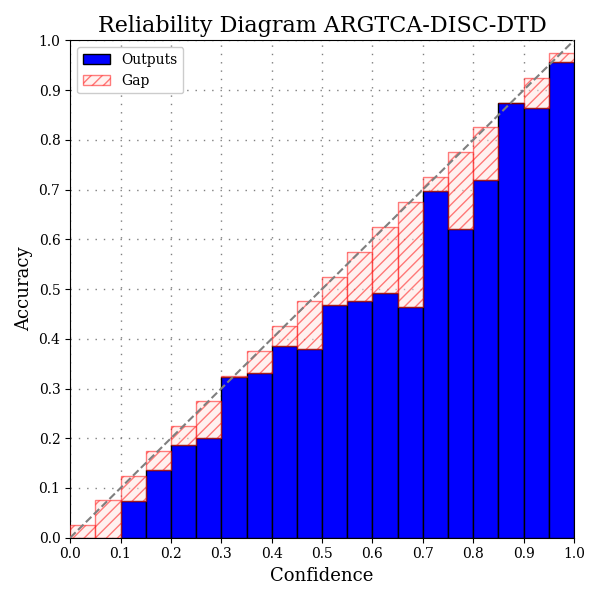} \\[3pt]

\multicolumn{6}{l}{\small\textit{Stanford Cars}} \\[0.5pt]
\includegraphics[width=\mw]{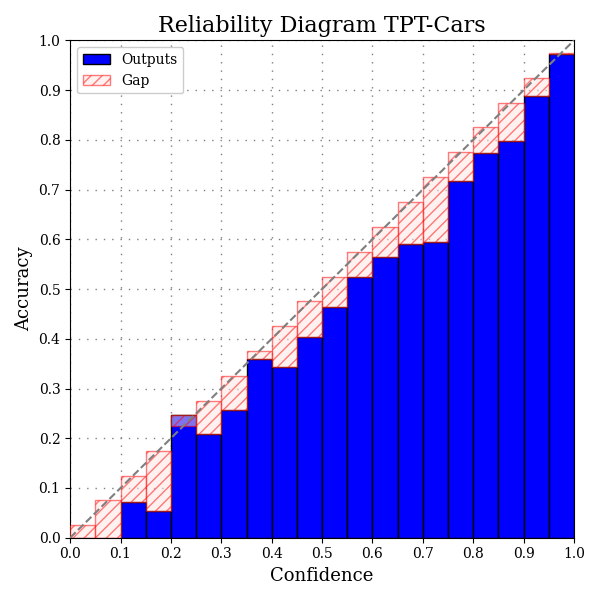} &
\includegraphics[width=\mw]{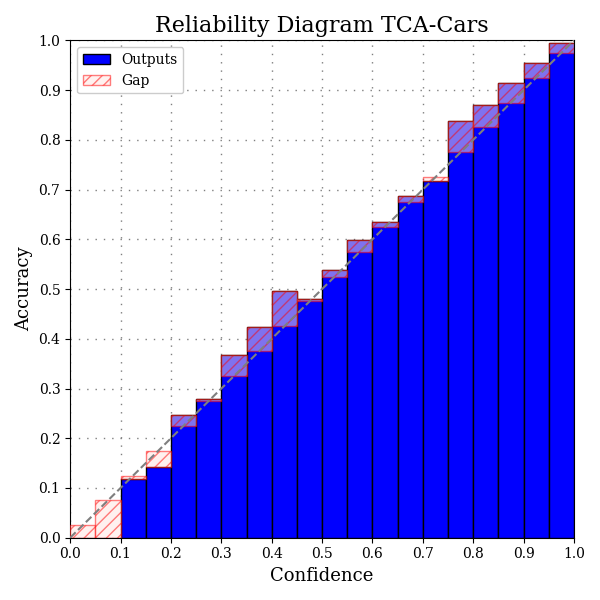} &
\includegraphics[width=\mw]{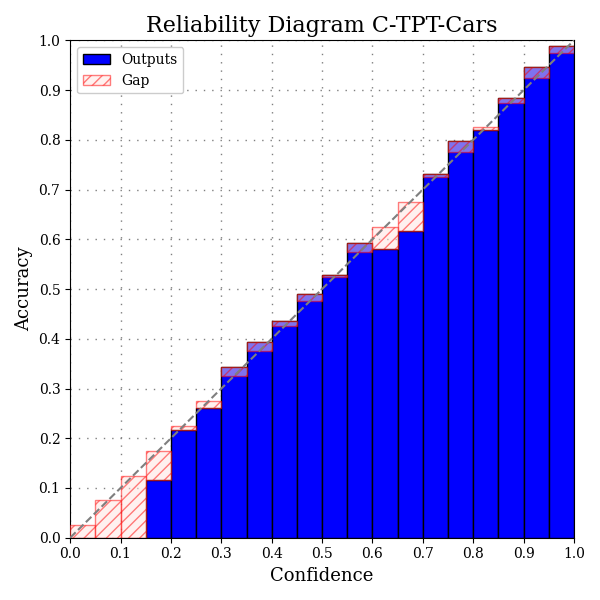} &
\includegraphics[width=\mw]{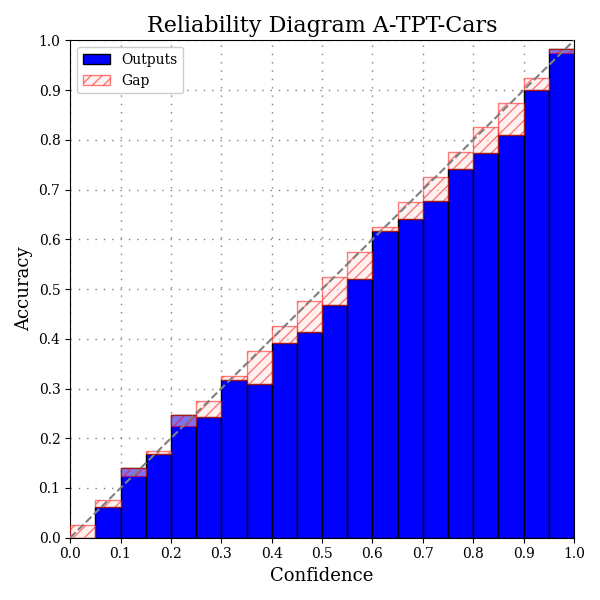} &
\includegraphics[width=\mw]{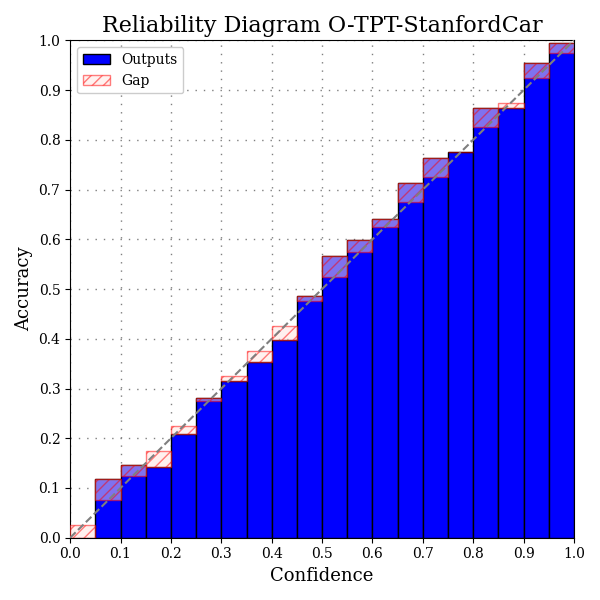} &
\includegraphics[width=\mw]{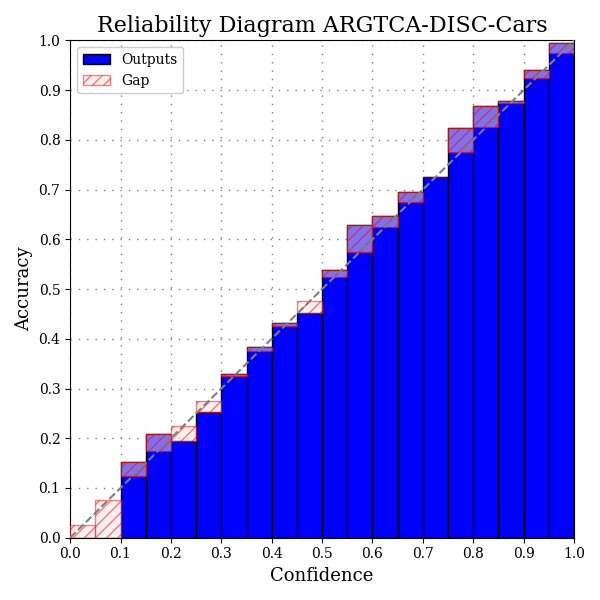} \\[3pt]

\multicolumn{6}{l}{\small\textit{Food101}} \\[0.5pt]
\includegraphics[width=\mw]{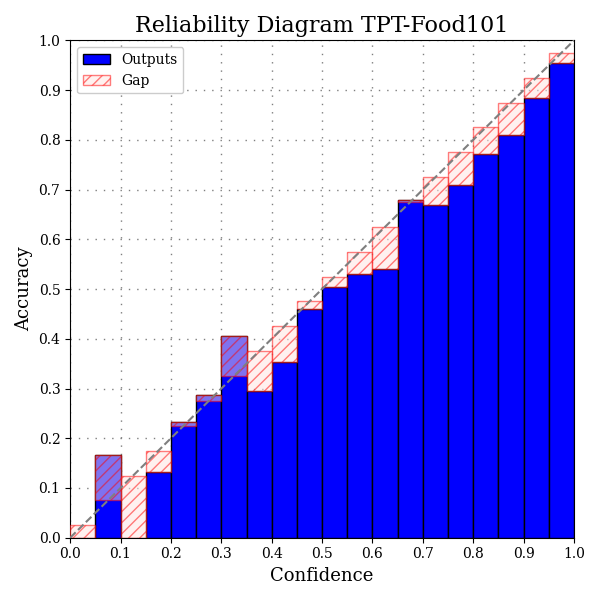} &
\includegraphics[width=\mw]{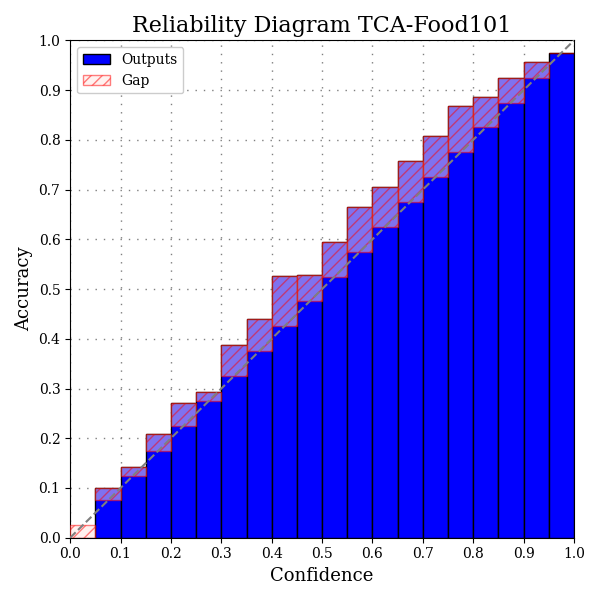} &
\includegraphics[width=\mw]{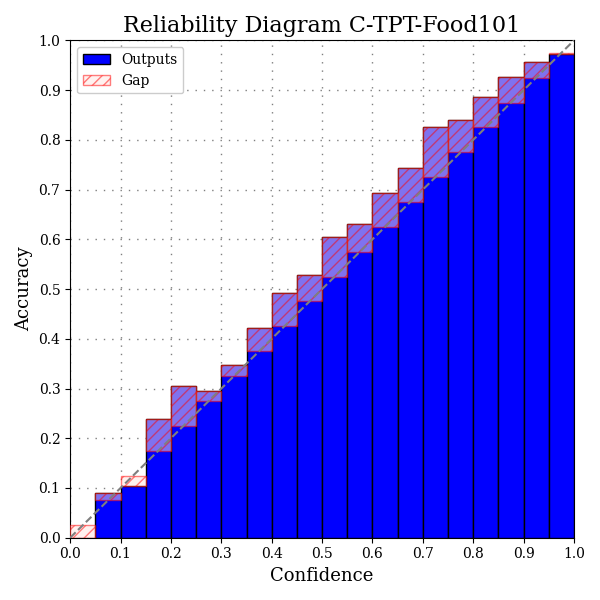} &
\includegraphics[width=\mw]{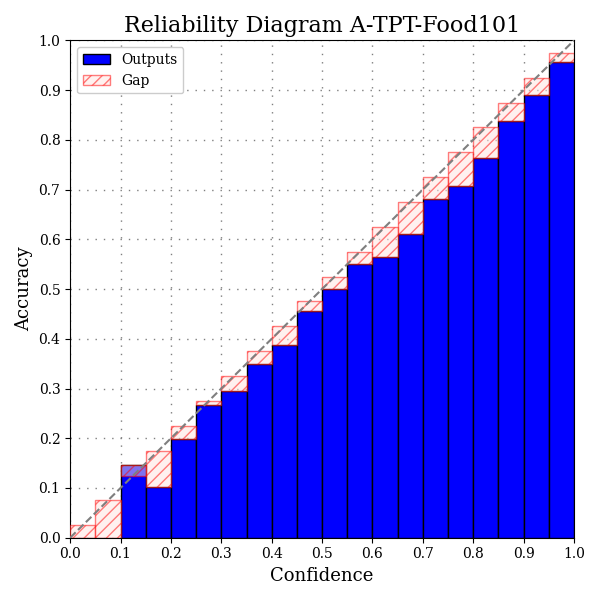} &
\includegraphics[width=\mw]{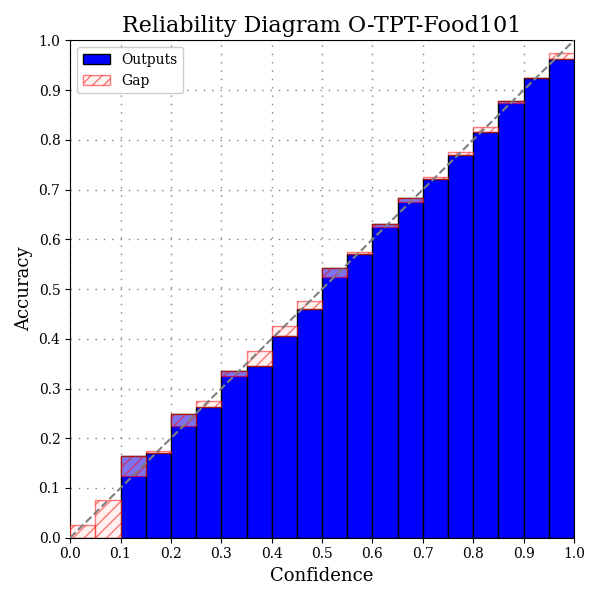} &
\includegraphics[width=\mw]{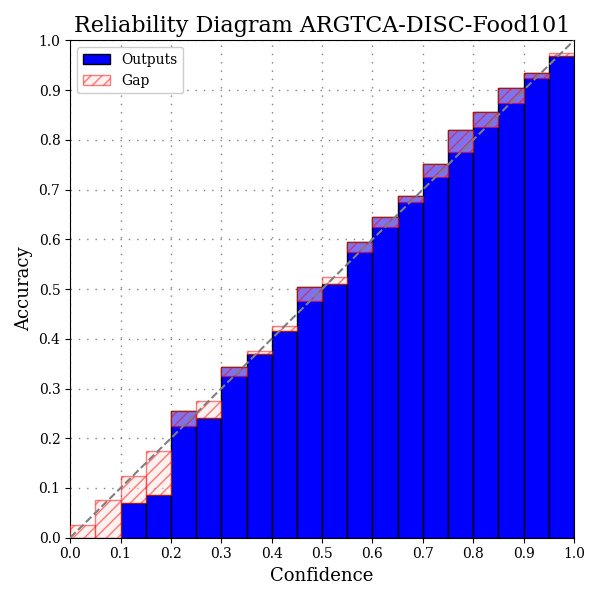} \\[3pt]

\multicolumn{6}{l}{\small\textit{Oxford Pets}} \\[0.5pt]
\includegraphics[width=\mw]{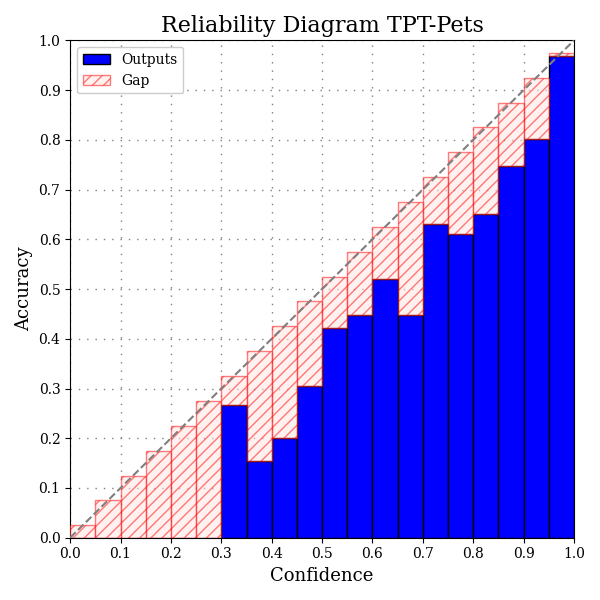} &
\includegraphics[width=\mw]{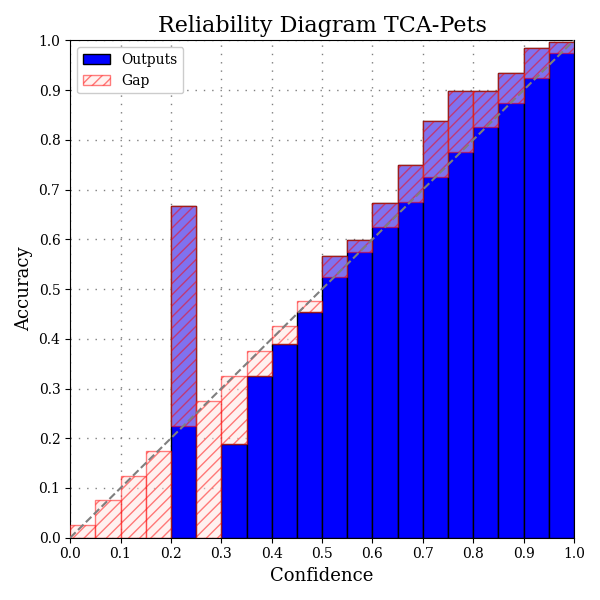} &
\includegraphics[width=\mw]{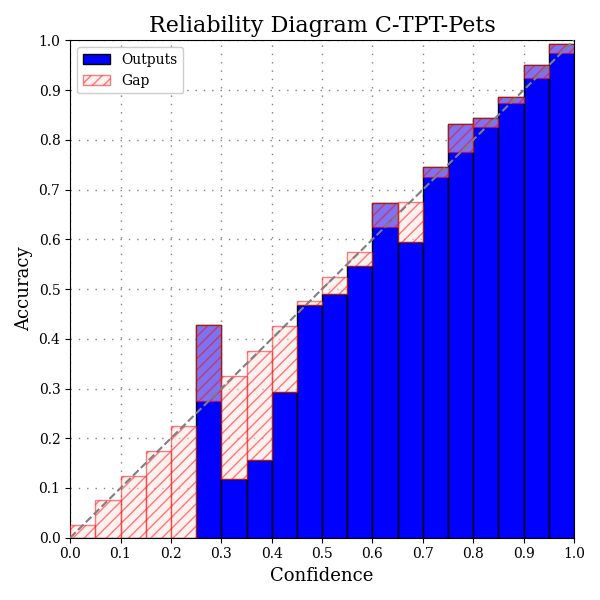} &
\includegraphics[width=\mw]{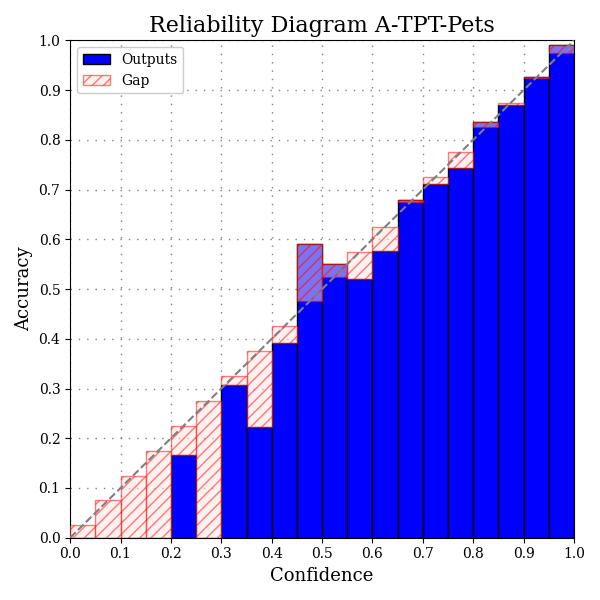} &
\includegraphics[width=\mw]{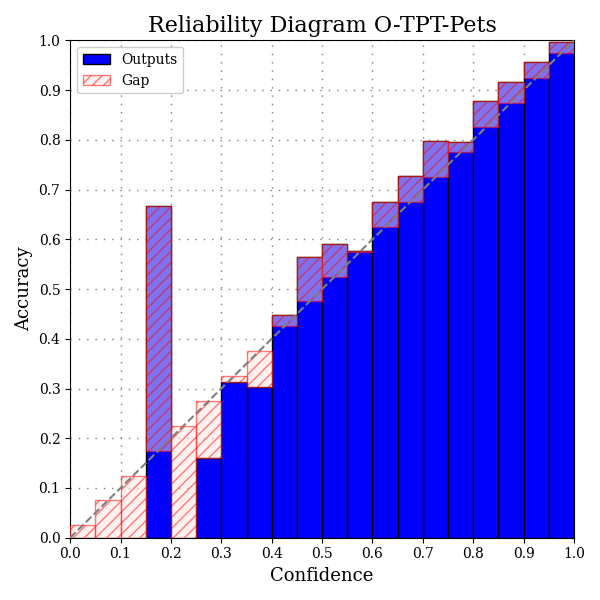} &
\includegraphics[width=\mw]{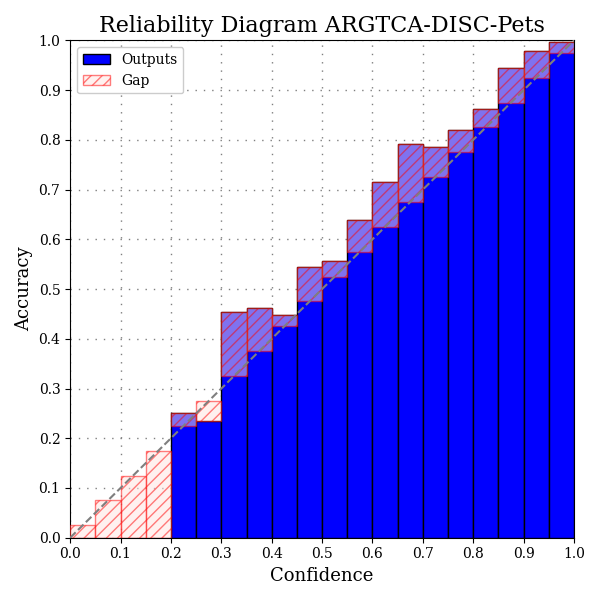} \\[3pt]

\multicolumn{6}{l}{\small\textit{EuroSAT}} \\[0.5pt]
\includegraphics[width=\mw]{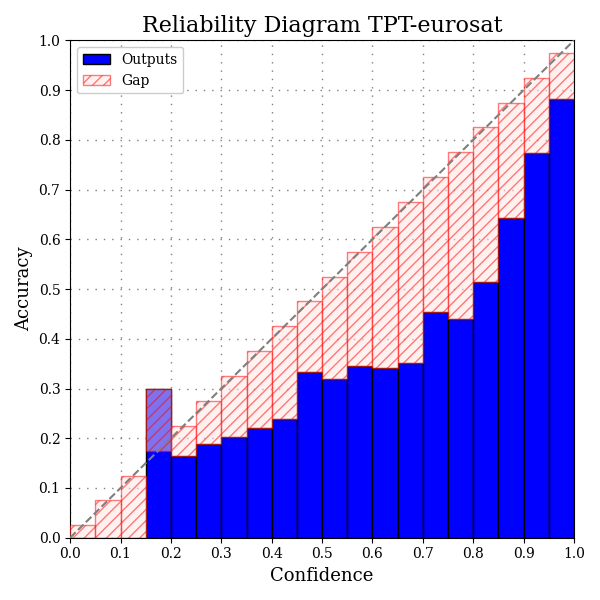} &
\includegraphics[width=\mw]{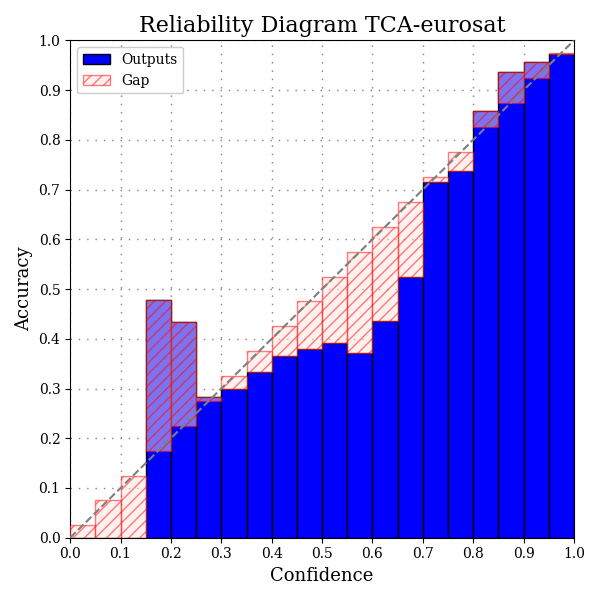} &
\includegraphics[width=\mw]{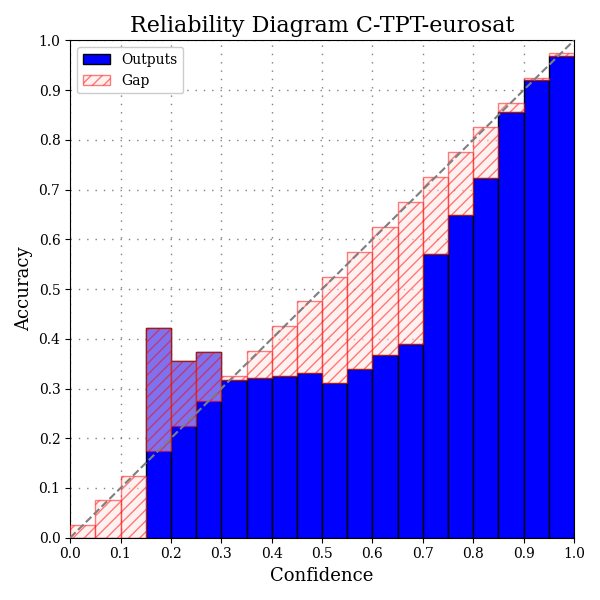} &
\includegraphics[width=\mw]{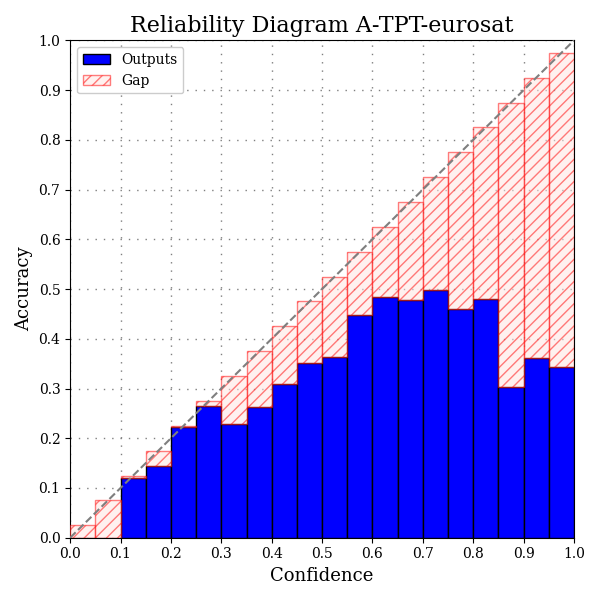} &
\includegraphics[width=\mw]{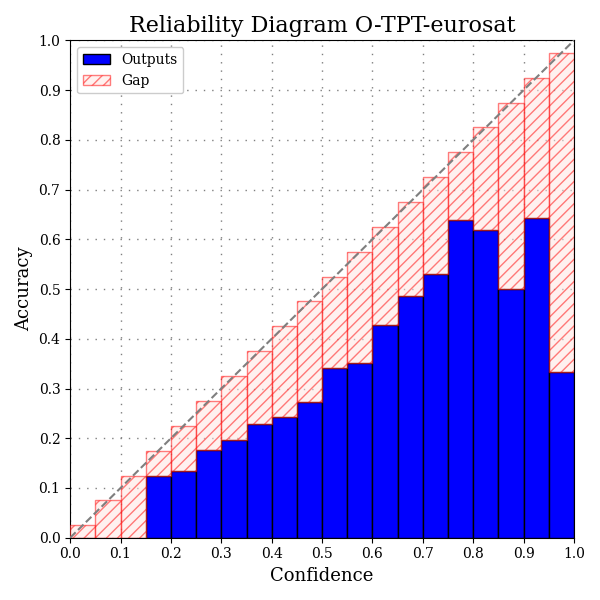} &
\includegraphics[width=\mw]{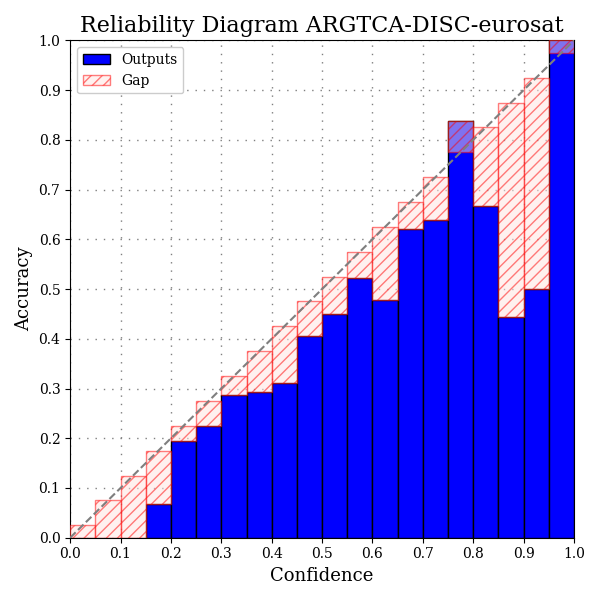} \\[3pt]

\multicolumn{6}{l}{\small\textit{Aircraft}} \\[0.5pt]
\includegraphics[width=\mw]{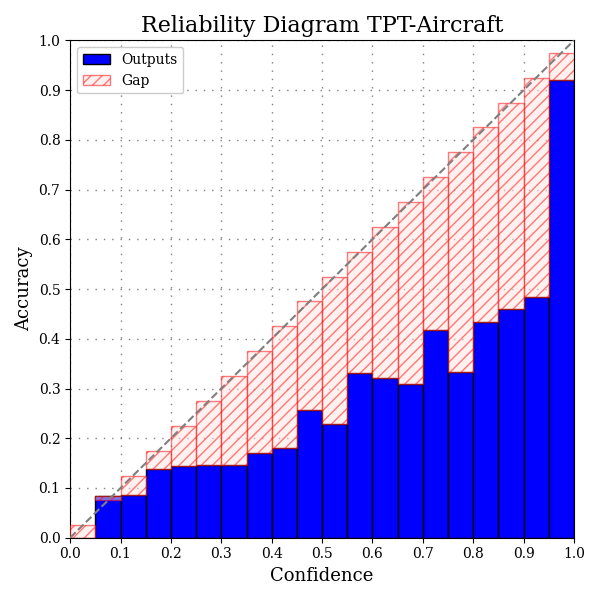} &
\includegraphics[width=\mw]{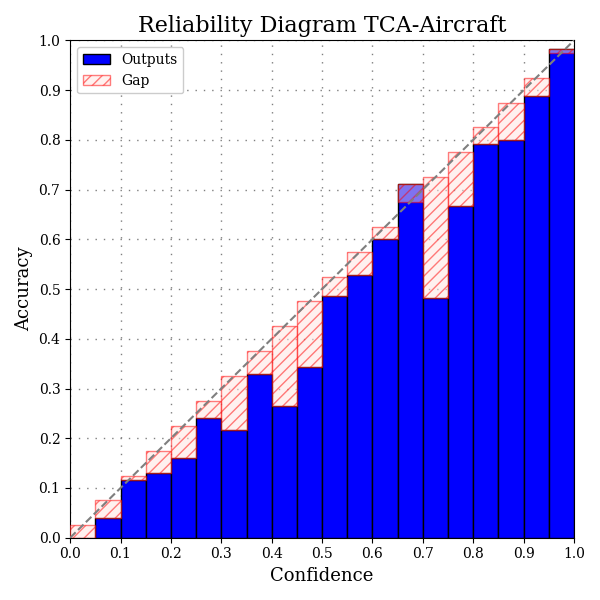} &
\includegraphics[width=\mw]{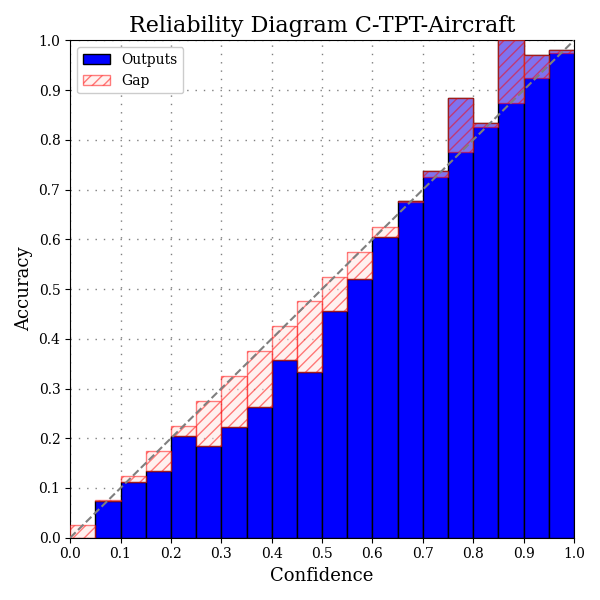} &
\includegraphics[width=\mw]{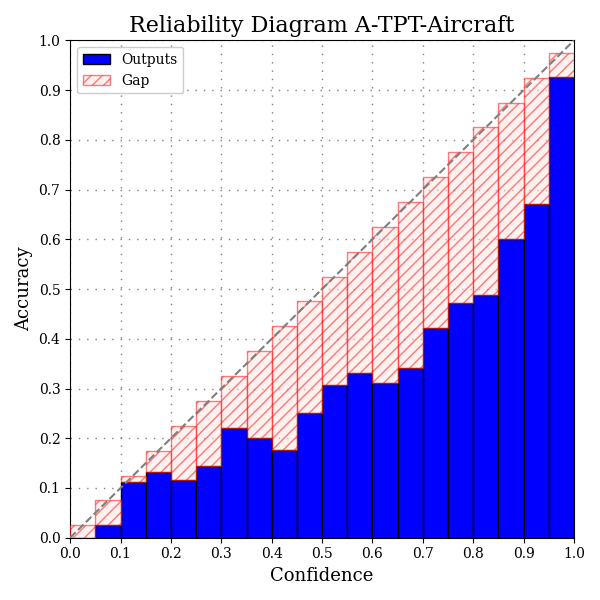} &
\includegraphics[width=\mw]{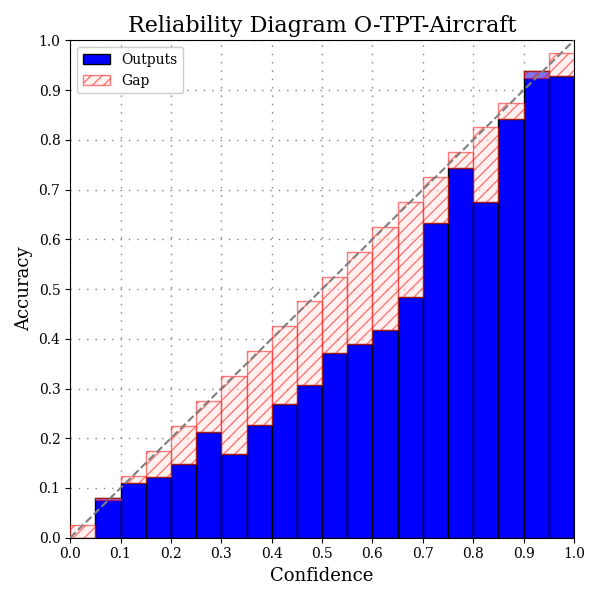} &
\includegraphics[width=\mw]{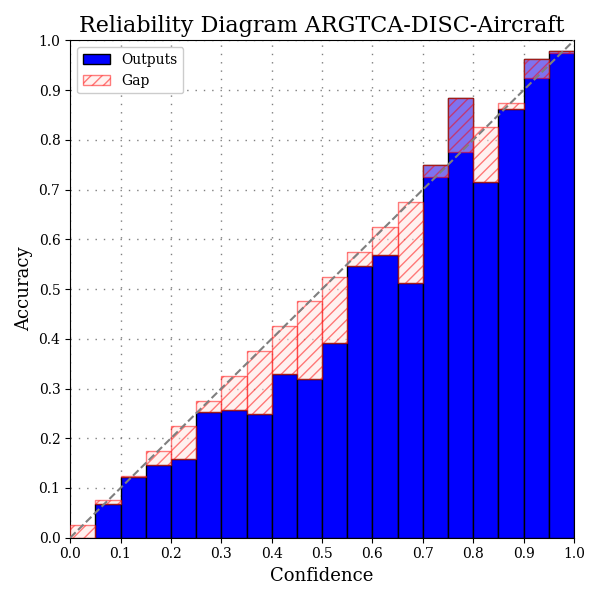} \\[3pt]

\multicolumn{6}{l}{\small\textit{Flower102}} \\[0.5pt]
\includegraphics[width=\mw]{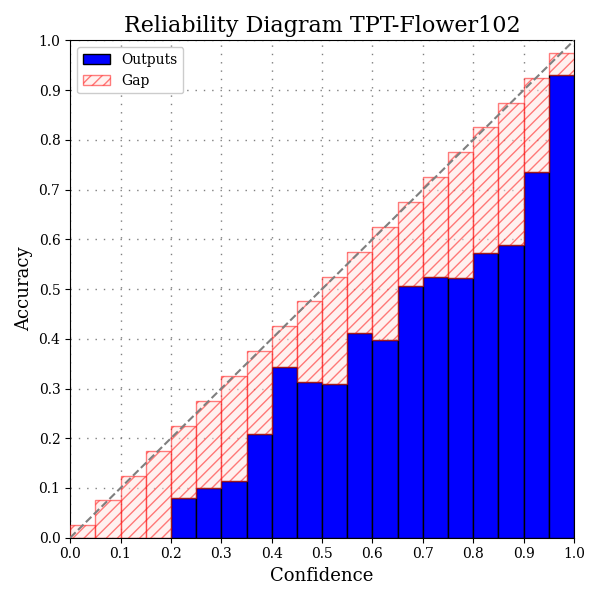} &
\includegraphics[width=\mw]{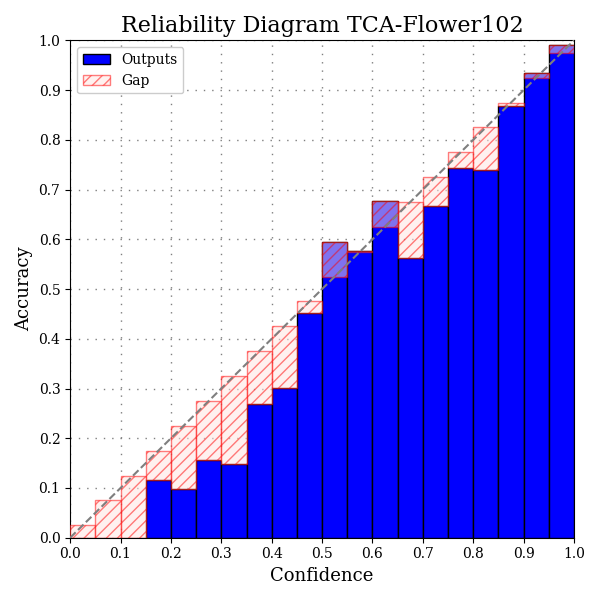} &
\includegraphics[width=\mw]{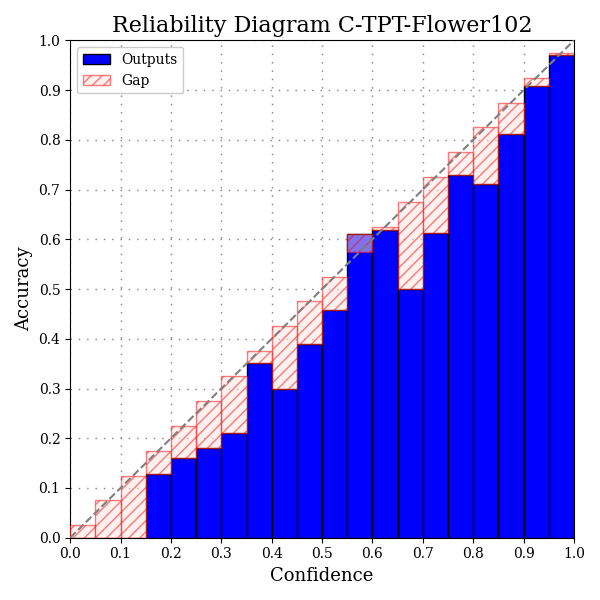} &
\includegraphics[width=\mw]{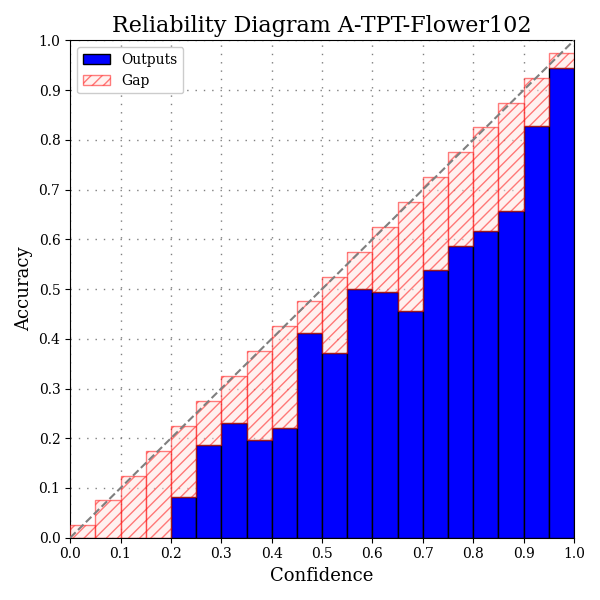} &
\includegraphics[width=\mw]{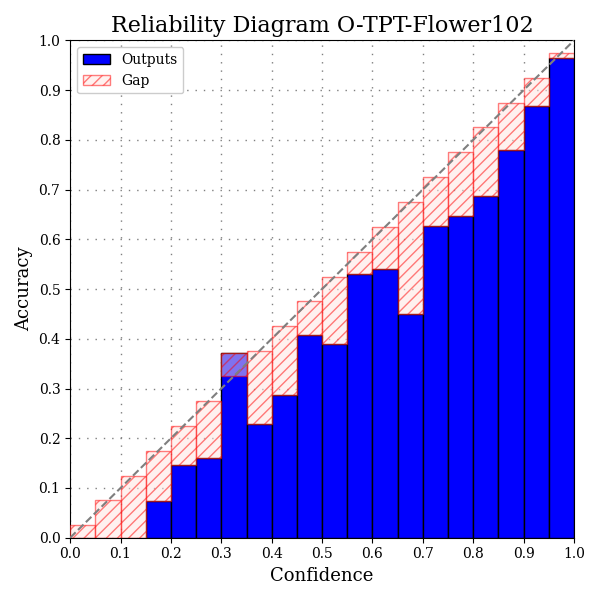} &
\includegraphics[width=\mw]{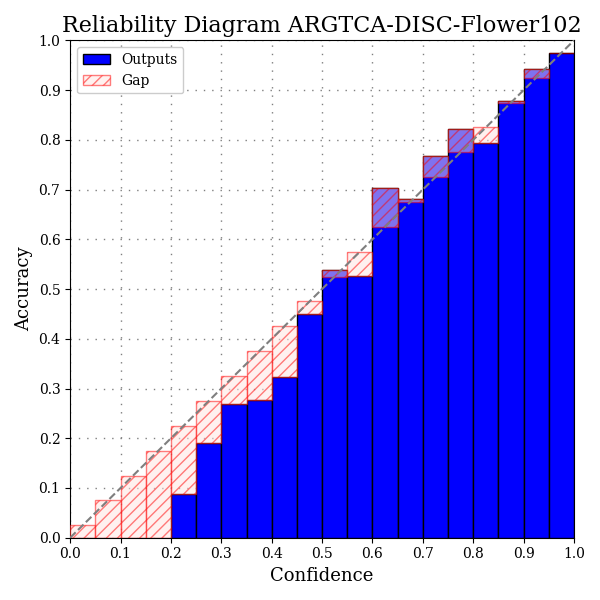} \\

\end{tabular}

\caption{Reliability diagrams across all datasets using ViT-B/16. Each row is a dataset; each column is a method. Bars above/below the diagonal (pink) indicate under/overconfidence. \ARGTCA{} consistently shows better alignment with the diagonal across all datasets.}
\label{fig:reliability-all}
\end{figure*}

\section*{Notation Reference}
A full list of all the notations used in the document are provided in Table~\ref{tab:notations}.
\begin{table*}[h]
\centering
\renewcommand{\arraystretch}{1.25}
\begin{tabular}{@{}lll@{}}
\toprule
\textbf{Symbol} & \textbf{Shape / Type} & \textbf{Meaning} \\
\midrule
$K$                          & scalar               & Number of classes \\
$M$                          & scalar               & Attributes loaded per class (\texttt{num\_attributes}$=10$) \\
$M'$                         & scalar               & Attributes selected for injection (\texttt{prompt\_attributes}$=2$) \\
$N = K \times M$             & scalar               & Total graph nodes \\
$d = 512$                    & scalar               & CLIP token embedding dimension \\
$c_k$                        & string               & Class name for class $k$ \\
$a_j$                        & string               & $j$-th attribute string \\
$(c_k, a_j)$                 & tuple                & Node $i$ in the graph \\
$\hzero_i$                   & $\R^d$               & Initial node embedding (CLIP EOS hidden state) \\
$\hL_i$                      & $\R^d$               & GAT output (refined node embedding) \\
$\p$                         & $\R^{n_{\text{ctx}} \times d}$ & Soft prompt context (trainable, Phase~2) \\
$g(\cdot)$                   & $\R^d$               & CLIP text encoder EOS hidden state \\
$f(\cdot)$                   & $\R^{d_{\text{out}}}$ & CLIP text encoder full output (after projection $\mathbf{T}$) \\
$\tau$                       & scalar               & InfoNCE temperature \\
$\alpha,\,\beta$             & scalars              & TCA inter / intra loss weights \\
\bottomrule
\end{tabular}
\caption{Notation used throughout this document.}
\label{tab:notations}
\end{table*}

\section{GAT Architecture Details}
\label{app:architecture}

Our graph-based attribute selection follows the Graph Attention Network (\GAT{})
framework of \citet{velickovic2018graph}, which assigns learned importance
weights to neighborhood edges during message passing, enabling the model
to selectively aggregate information from the most relevant neighbors. Our \SAG{} is an \textit{undirected} graph; each undirected edge
$\{u, v\} \in \mathcal{E}$ is represented as two directed edges
$(u \to v)$ and $(v \to u)$ during message passing, following standard
practice in \GAT{} implementations.

\paragraph{Attention and aggregation.}
For each edge $\{u,v\}$ and head $h$, the normalized attention coefficient is:
\begin{equation}
\small
  \alpha_{uv,h} = \frac{
    \exp\!\left(\mathrm{LeakyReLU}(
      \mathbf{a}_h^\top[\widetilde{\mathbf{h}}_{u,h} \| \widetilde{\mathbf{h}}_{v,h}])\right)
  }{
    \sum_{u'\in\mathcal{N}(v)}
    \exp\!\left(\mathrm{LeakyReLU}(
      \mathbf{a}_h^\top[\widetilde{\mathbf{h}}_{u',h} \| \widetilde{\mathbf{h}}_{v,h}])\right)
  },
\end{equation}
where $\widetilde{\mathbf{h}} = \mathbf{h}^{(l)}W_{\text{layer}}$ is the
linearly projected embedding. Node representations are updated as:
\begin{equation}
\small
  \widetilde{\mathbf{h}}^{(l)}
  = \mathrm{ELU}\!\left(
      \mathrm{concat}_{h=1}^{H}
      \!\left[\sum_{u\in\mathcal{N}(v)}
      \mathrm{drop}(\alpha_{uv,h})\cdot\widetilde{\mathbf{h}}_{u,h}\right]
    \right).
\end{equation}

\paragraph{Residual, normalisation, and projection.}
After $L$ layers, a residual connection (with $W_{\mathrm{res}}$
initialised to $\mathbf{I}$) mitigates over-smoothing,
followed by $\ell_2$ normalisation:
\begin{equation}
  \mathbf{h}^{(L)}
  = \mathrm{normalize}\!\left(
      \widetilde{\mathbf{h}}^{(L)} + W_{\mathrm{res}}\,\mathbf{h}^{(0)}
    \right).
\end{equation}
A two-layer MLP projects to the final embedding:
$\mathbf{z} = W_2(\mathrm{GELU}(W_1\mathbf{h}^{(L)}+\mathbf{b}_1))$.

\paragraph{Node attention score.}
A scalar importance score per node, used for attribute selection
in \ARGTCADiv{} and \ARGTCADisc{}, is computed as the mean attention
coefficient across heads and layers:
\begin{equation}
  \mathrm{attn}[v]
  = \frac{1}{LH} \sum_{l=1}^{L}\sum_{h=1}^{H}
    \frac{1}{|\mathcal{N}(v)|}
    \sum_{u\in\mathcal{N}(v)} \alpha_{uv,h}.
\end{equation}
This scalar serves as the basis for attribute selection in both
\ARGTCADiv{} and \ARGTCADisc{} (Section~\ref{sec:method:phase1}).

\section{Ablation Studies}
\label{app:ablation}

\subsection{Attribute comparison: TCA vs.\ \ARGTCADiv{} vs.\ \ARGTCADisc{}.}
The table~\ref{tab:attr_comparison} lists the two attributes selected per method for a representative subset of classes from Caltech101, OxfordPets, and DTD. \TCA{} ranks attributes by cosine similarity between each attribute's frozen \CLIP{} embedding and the class name embedding, selecting the top-$M'$; \ARGTCADiv{} selects the most complementary pair by minimum pairwise cosine similarity in $\hL$ space; \ARGTCADisc{} selects the pair with highest mean cosine distance to all other-class embeddings in $\hL$ space.

\subsection{Zero graph edges.}
To isolate the contribution of the \SAG{} edge structure, we train the \GAT{} with $\calE = \emptyset$: all nodes are present and trained via supervised contrastive loss, but no message passing occurs. Without edges, each \GAT{} layer reduces to an independent per-node linear transform; no cross-class or within-class structural context propagates. Attribute selection uses the same Div/Disc criteria on these edge-free embeddings. The calibration drop on DTD relative to the full model (Table~\ref{tab:ablation}) confirms that the \SAG{}'s edge topology --- not the contrastive objective alone --- drives selection quality.

\subsection{No \GAT{} training (raw \CLIP{} EOS embeddings).}
We skip Phase~1 entirely and apply the Div/Disc selection criteria directly to the initial embeddings $\hzero$ (frozen \CLIP{} EOS hidden states, Eq.~\ref{eq:h0}), without any contrastive training. This tests whether \GAT{} training adds value beyond the semantic geometry already present in \CLIP{}'s encoder. The DTD ECE degrades substantially (Disc: $8.95$ vs.\ $6.36$ for the full model), showing that contrastive relational training is necessary for reliable attribute discrimination on texture-heavy datasets.

\subsection{Random attribute selection.}
We run Phase~1 in full (SAG construction, \GAT{} training) but replace Div/Disc selection with a random draw of $M' = 2$ attributes per class (seed 42). This isolates the selection criterion: random selection degrades ECE on DTD (Div: $5.16$ vs.\ $3.17$ for the full model) while leaving Caltech accuracy largely unchanged, confirming that the geometric Div/Disc criteria are the operative mechanism for calibration improvement, not merely the graph training.

\subsection{Hyperparameter sensitivity ($\alpha$, $\beta$).}
Figures~\ref{fig:alpha} and~\ref{fig:beta} show ECE as $\alpha$ and $\beta$ are swept independently ($\alpha \in \{5,10,20,40\}$ with $\beta{=}35$ fixed; $\beta \in \{15,25,35,50\}$ with $\alpha{=}10$ fixed), evaluated on Caltech101, OxfordPets, and DTD for both \ARGTCADiv{} and \ARGTCADisc{}. Both metrics are stable across the swept range: ECE varies by at most ${\sim}1.5\%$ on Caltech101 and ${\sim}2.0\%$ on DTD for $\alpha$, and by at most ${\sim}1.8\%$ on DTD for $\beta$. The optimal region is $\alpha \in [10,20]$ and $\beta \in [25,35]$; the ViT-B/16 defaults ($\alpha{=}10$, $\beta{=}35$) lie within this plateau, confirming robustness to precise hyperparameter choice.

\begin{figure}[h]
    \centering
     \includegraphics[width=1\linewidth]{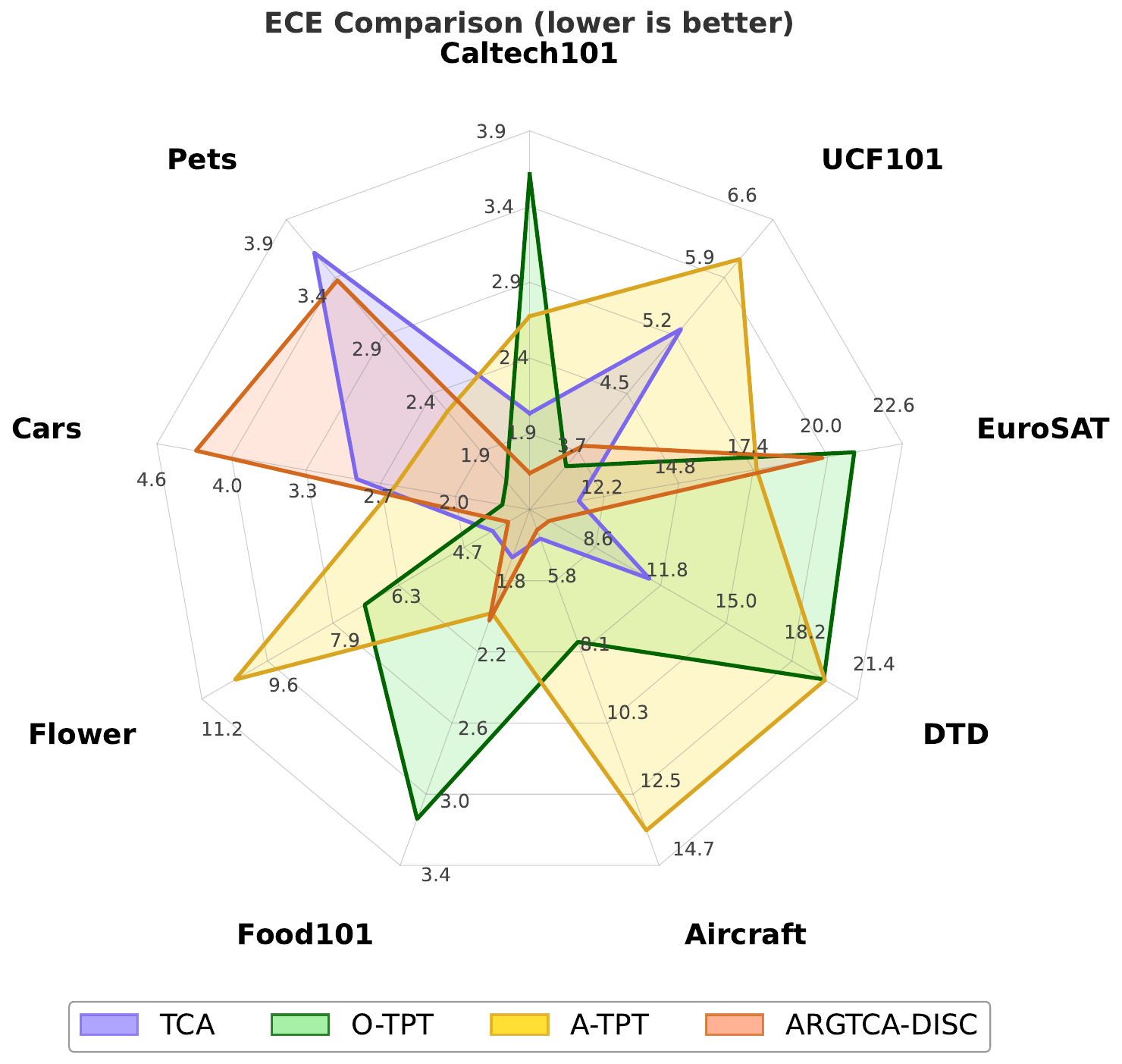}
    \caption{Radar plot comparing \ECE{} (lower is better, smaller area is better) of \TCA{}, \OTPT{}, A-TPT, and \ARGTCADisc{} (Ours) across nine datasets (ViT-B/16).}
    \label{fig:radar_disc}
\end{figure}
\end{document}